\newcommand{\shortversion}[1]{}
\newcommand{\longversion}[1]{#1}

\documentclass[conference]{IEEEtran}
\usepackage{times}

\usepackage[numbers]{natbib}
\usepackage{multicol}
\usepackage[bookmarks=true]{hyperref}

\usepackage[fleqn]{amsmath}
\usepackage{xspace}
\usepackage{float}
\usepackage{balance}
\usepackage{amssymb,latexsym,amsfonts,amsmath,amsthm}
\usepackage{xkeyval}
\usepackage{color}
\usepackage{tikz}
\usetikzlibrary{automata}
\usetikzlibrary{shapes,arrows}
\usepackage[latin1]{inputenc}
\usepackage{dcolumn}
\usepackage{multirow}
\usepackage{multicol}
\usepackage{subfigure}
\usepackage{eso-pic}
\usepackage{epsfig}
\usepackage{url}
\usepackage{xcolor}
\usepackage{pgfplots}

\usepackage{tikz}
\usepackage{siunitx}

\usepackage[algo2e,ruled,vlined,linesnumbered]{algorithm2e}
\usepackage[nodisplayskipstretch]{setspace}
\SetAlFnt{\small\rmfamily}
\usetikzlibrary{patterns}
\usepgfplotslibrary{external}
\usepackage{subfigure}
\usepackage{array}
\newcolumntype{P}[1]{>{\arraybackslash}p{#1}}
\usepackage{multicol}
\usepackage{etoolbox}

\DeclareMathOperator*{\argmin}{argmin}

\newtheorem{lemma}{{\sc Lemma}}

\newtheorem{theorem}{{Theorem}}

\newtheorem{example}{{\sc Example}}
\AtEndEnvironment{example}{\qed \vspace{-1mm}}

\newtheorem{question}{{\sc Question}}
\newtheorem{problem}{{\sc Problem}}

\begin{document}

\title{\textsf{MT*} : Multi-Robot Path Planning for Temporal Logic Specifications}

\author{Author Names Omitted for Anonymous Review. Paper-ID 252}

\author{\authorblockN{Dhaval Gujarathi}
\authorblockA{SAP Labs, India\\
Email: dhavalsgujarathi@gmail.com}
\and
\authorblockN{Indranil Saha}
\authorblockA{IIT Kanpur, India\\
Email: isaha@cse.iitk.ac.in}
}



%

\maketitle

\begin{abstract}
We address the path planning problem for a team of robots satisfying a complex high-level mission specification given in the form of an Linear Temporal Logic (LTL) formula. 
The state-of-the-art approach to this problem employs the automata-theoretic model checking technique to solve this problem. This approach involves computation of a product graph of the B\"uchi automaton generated from the LTL specification and a joint transition system which captures the collective motion of the robots and then computation of the shortest path using Dijkstra's shortest path algorithm. We propose \textsf{MT*}, an algorithm that reduces the computation burden for generating such plans for multi-robot systems significantly. Our approach generates a reduced version of the product graph without computing the complete joint transition system, which is computationally expensive. It then divides the complete mission specification among the participating robots and generates the trajectories for the individual robots independently. Our approach demonstrates substantial speedup in terms of computation time over the state-of-the-art approach, and unlike the state of the art approach, scales well with both the number of robots and the size of the workspace.
\end{abstract}

\IEEEpeerreviewmaketitle

\section{Introduction} 
\label{sec-introduction} 
Path planning is one of the core problems in robotics, where we design algorithms to enable autonomous robots to carry out a real-world complex task successfully~\cite{LaValle:2006:PA:1213331}. A basic path planning task involves point-to-point navigation while avoiding obstacles and satisfying some user-given constraints. 
Recently, there has been an increased interest in specifying complex paths for the robots using temporal logic (e.g.~\cite{Kress-GazitFP07,KaramanF09,BhatiaKV10,sampling5509503,WongpiromsarnTM12,ChenTB12,UlusoySDBR13,SahaRKPS14,KhalidiGS20}).
Using temporal logic~\cite{Baier:2008:PMC:1373322}, one can specify requirements that involve a temporal relationship between different operations performed by robots.

This paper focuses on the class of multi-robot Linear Temporal Logic (LTL) path planning problems where the inputs are the \emph{discrete dynamics} of the robots and a global LTL specification.
Though we deal with any specification that  can be captured in LTL, 
our main focus is to deal with those specifications that require the robots to repeat some tasks perpetually. 
Such requirements arise in many robotic applications, including persistent surveillance~\cite{UlusoySDBR13}, assembly planning~\cite{Halperin1998}, 
evacuation~\cite{RodriguezA10}, search and rescue~\cite{Jennings97}, localization~\cite{Fox00}, object transportation~\cite{Rus95}, and formation control~\cite{Balch98}.

Traditionally, the LTL path planning problem for the robots with discrete dynamics is reduced to the problem of finding the shortest path in a weighted graph, and Dijkstra's shortest path algorithm~\cite{Cormen:2009:IAT:1614191} is employed to generate an optimal trajectory satisfying an LTL query~\cite{belta5650896,UlusoySDBR13}. 
However, for a large workspace and a complex LTL specification, this approach is merely scalable.
We seek to design a computationally efficient algorithm to generate optimal trajectories for the robots.  

Heuristic based search algorithms such as A* (for a single robot)~\cite{Russell:2009:AIM:1671238} and M* (for a multi-robot system)~\cite{WAGNER20151} have been successfully applied to solving point-to-point path planning problems and are proven to be significantly faster than Dijkstra's shortest path algorithm. Heuristic search based algorithms have also been applied to solving the temporal logic path planning problem for a single robot~\cite{BacchusK98,BaierM06a,BaierM06b,PatriziLGG11,KhalidiGS20}.
In this paper, we introduce the \textsf{MT*} algorithm that, for the first time, attempts to incorporate the heuristic search in 
generating an \emph{optimal} trajectory for a multi-robot system satisfying an global LTL query efficiently. 
We apply our algorithm to solving various LTL path planning problems for a multi-robot system in 2-D workspaces and compare the results with that of the algorithm presented in~\cite{UlusoySDBR13}.
Our experimental results demonstrate that \textsf{MT*} in many cases achieves an order of magnitude better computation time than that of the traditional approach~\cite{UlusoySDBR13} to solve the optimal LTL path planning problem.

\smallskip
\noindent
\textbf{Related Work.} \textsf{MT*} is a special class of multi-agent path finding (MAPF) problem.
MAPF is a widely studied planning problem where the goal is to find collision-free paths for a number of agents from their initial locations to some specified goal locations~\cite{MaK17,SahaRKPS16}. MAPF is a special class of the finite LTL path synthesis problem~\cite{CamachoBMM18} for multi-agent systems where the specification of each robot is given by the LTL formula ``eventually $goal$'', where $goal$ is the proposition that becomes $\mathtt{true}$ when all the robots reach their goal locations. 
A number of previous works addressed the multi-robot path planning problem for general finite LTL specifications~\cite{SahaRKPS14,SchillingerBD16,SchillingerBD17}.

On the contrary, our focus in the paper is to address the planning problem for those LTL specifications that capture perpetual behavior and is satisfied using infinite trajectories, like the work presented in~\cite{Kloetzer10, Chen12, UlusoySDBR13,TumovaD16,Shoukry17,KantarosZ19,KantarosZ20}.
Among the above-mentioned work, Kloetzer et al.~\cite{Kloetzer10} and Shoukry et al.~\cite{Shoukry17} solve the problem without discretizing the robot dynamics. However, such techniques are computationally demanding and scale poorly with the number of robots. In all the other papers, the robot dynamics are discretized into weighted transition systems.
The work by Tumova et al.~\cite{TumovaD16} assumes that the robots are assigned their own tasks, with some communication requirement among the robots.
In this paper, we deal with a single specification for the multi-robot system, where finding the optimal distribution of the tasks is the core challenge.
Chen et al.~\cite{Chen12} divide the problem by first decomposing the given specification into the specifications for the individual robots and then generating the plans for the robots based on their own specification. This decoupling leads to suboptimality as a given LTL specification may have a number of valid decompositions.
and their quality can be known only after the plans are generated.
Recent work by Kantaros and Zavlanos to solve the LTL path planning problem scales for a large number of robots~\cite{KantarosZ19,KantarosZ20}. 
They employ sampling based technique to compute the first feasible trajectory, which might not be cost-optimal. Unlike their work, we focus on generating trajectories with minimum cost. 


\section{Problem}
\label{sec-problem}

\subsection{Preliminaries} 
\label{subsec-preliminaries}

\subsubsection{Workspace, Robot Actions and Trajectory}
We assume that a team of $n$ robots operates in a 2-D or a 3-D discrete workspace $\mathcal{W}$ which we represent as a grid map. The grid divides the workspace into square-shaped cells. Every cell in the workspace $\mathcal{W}$ is referenced using its coordinates. Some cells can be marked as obstacles and cannot be visited by any robot. We denote the set of obstacles using $\mathcal{O}$.
We capture the motion of a robot using a set of actions $Act$. The robot changes its state in the workspace by performing the actions from $Act$. An action $act \in Act$ is associated with a $cost$, which captures the energy consumption or time delay (based on the need) to execute it. A robot can move to satisfy a given specification by executing a sequence of actions in $Act$ generating a \emph{trajectory} of states it attains. The \emph{cost of a trajectory} is the sum of costs of the actions to generate the trajectory. 

\subsubsection{Transition System}
We model the motion of the robot $i$ in the workspace $\mathcal{W}$ as a weighted transition system defined as
\mbox{$T^i := (S^i, s_0^i, E^i,\Pi^i, L^i, w^i),$}
\noindent where (i) $S^i$ is the set of states,  
(ii) $s_0^i \in S^i$ is the initial state of the robot $i$,  (iii) $E^i \subseteq S^i \times S^i$ is the set of transitions/edges allowed to be taken by robot $i$, $(s_1^i , s_2^i) \in E^i$  iff  $s_1^i , s_2^i \in S^i$ and $s_1^i$ $\xrightarrow[]{act}$ $s_2^i$, where $act \in Act$,  (iv) $\Pi^i$ is the set of atomic propositions defined for robot $i$, (v) $L^i : S^i$ $\rightarrow$ $2^{\Pi^i}$ is a map which provides the set of atomic propositions satisfied at a state, (vi) $w^i: E^i$ $\rightarrow$ $\mathbb{N}_{>0}$ is a weight function.


\subsubsection{Joint Transition System}
A joint transition system $T$ is a transition system that captures the collective motion of a team of $n$ robots in a workspace$(\mathcal{W})$, where each robot executes one action from the set of actions $Act$ available to it. We define a joint transition system as
\mbox{$ T :=\: ( S_T,\; s_0,\; E_T,\; \Pi_T,\; L_T,\; w_T ),$}
where (i) $S_T$ is the set of vertices/states in a joint transition system, where each vertex is of form $\langle s^1, s^2,...,s^n\rangle$, $s^i$ represents the state of robot $i$ in transition system $T^i$, (ii) $s_0:=\langle s^1_0, s^2_0,...,s^n_0\rangle  \in S_T$ is the joint initial state of the team of $n$ robots, (iii) $E_T \subseteq S_T \times S_T$ is the set of edges. $(s_1,s_2) \in E_T$ iff $s_1 , s_2 \in S_T$ and $(s_1^i,s_2^i) \in E^i$ for all $i \in \{ 1,2,...,n\}$, (iv) $\Pi^T := \bigcup_{i=1}^n \Pi^{i}$ is the set of atomic propositions, (v) $L_T :  S_T \rightarrow 2^{\Pi_T}$, and $L_T(s_{1}) := \bigcup_{i=1}^n L^i(s_{1}^i)$ gives us set of propositions true at state $s_1$, (vi) $w_T : E^T \rightarrow \mathbb{N}_{>0}$, and $w_T( s_{1}, s_{2} ) := \sum_{i=1}^n w^i(s_{1}^i, s_{2}^i)$ is a weight function.

We can also think of the transition system as a weighted directed graph with vertices, edges, and a weight function. Whenever we use some graph algorithm over a transition system, we mean to apply it over its equivalent graph.

\begin{figure}[t] 
\centering
\subfigure[Workspace $\mathcal{W}$ with propositions $P_{1}, P_2 \text{ and } P_{3} $]{\centering \label{exampleGrid}\includegraphics[ width=40mm]{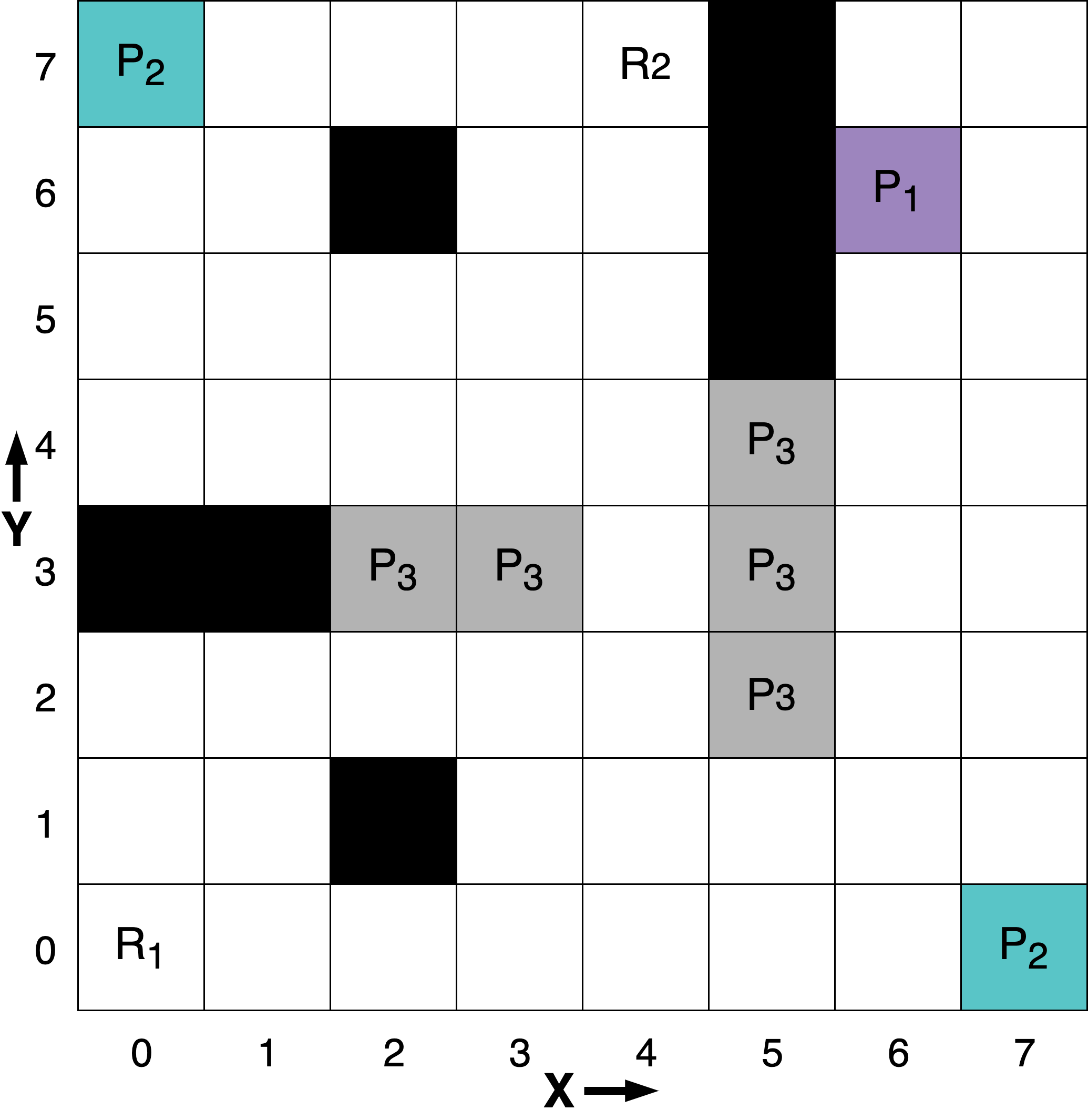}}
\hspace{2mm}
\subfigure[B\"{u}chi automaton B for query: $\square(\Diamond P_{1} \land \Diamond P_{2} \land \lnot P_{3})$]{\centering \label{exampleAutomata}\includegraphics[width=37mm]{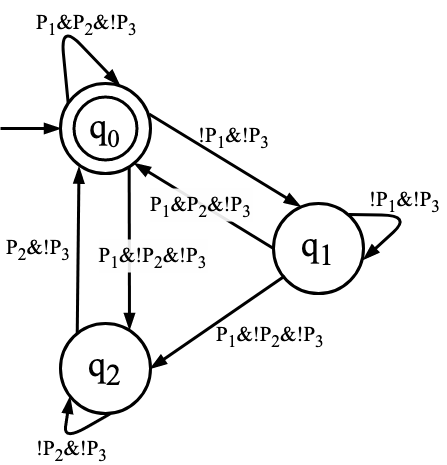}}
\caption{Workspace $\mathcal{W}$ and B\"uchi Automaton}
\label{ws}
\end{figure}

\begin{example}

Throughout this paper, we will use a warehouse pick-and-drop example for illustration purposes.
The workspace $\mathcal{W}$ is shown in Figure \ref{exampleGrid}. 
We build a transition systems $T^i$ for all the robots over $\mathcal{W}$ where $\Pi^i = \Pi_T = \{ P_1, P_2 , P_3 \}$. The proposition $P_i$ is satisfied if the robot is at one of the locations denoted by $P_i$.
Here, $P_1$, $P_2$, and $P_3$ correspond to drop location, pickup location, and the locations to be avoided by the robots, respectively. Cells with black colour represent obstacles ($\mathcal{O}$).
We assume that from any cell in $\mathcal{W}$, a robot can move to one of its four neighbouring cells with cost $1$ or stay at the same location with cost $0$. 
\end{example}

\subsubsection{Linear Temporal Logic}
The path planning query/task in our work is given in terms of formulae written using \emph{Linear Temporal Logic} (LTL).
\sloppy{LTL formulae over the set of atomic propositions $\Pi_T$ are formed according to the following grammar~\cite{Baier:2008:PMC:1373322}:
$$\Phi:: = \mathtt{true} \;|\; a \;| \;\phi_{1} \land \phi_{2} \;|\; \neg\phi \;|\; \text{X} \phi \;|\; \phi_{1} \, \text{U}\, \phi_{2} $$ 
The basic ingredients of an LTL formula are the atomic propositions $a \in \Pi_T$, the Boolean connectors like conjunction $\land$ and negation $\neg$, and two temporal operators $X$ (next) and $U$ (until). 
The semantics of an LTL formula is defined over an infinite trajectory $\sigma$. 
The trajectory $\sigma$ satisfies a formula $\xi$, if the first state of $\sigma$ satisfies $\xi$.
The logical operators conjunction $\land$ and negation $\neg$ have their usual meaning.
For an LTL formula $\phi$,
$X\phi$ is $\mathtt{true}$ in a state if $\phi$ is satisfied at the next step. The formula $\phi_1 \, \text{U} \, \phi_2$ denotes that $\phi_1$ must remain $\textsf{true}$ until $\phi_2$ becomes $\textsf{true}$ at some point in future.
The other LTL operators that can be derived are  $\square$ (\emph{Always}) and $\Diamond$ (\emph{Eventually}). The formula $\square \phi$ denotes that $\phi$ must be satisfied all the time in the future. The formula $\Diamond \phi$ denotes that $\phi$ has to hold sometime in the future. We have denoted negation $\lnot P$ as $!P$ and conjunction as $\&$ in the Figures. }

\subsubsection{B{\"u}chi Automaton}
For any LTL formula $\phi$ over a set of propositions $\Pi_T$, we can construct a B{\"u}chi automaton with input alphabet $\Pi_B = 2^{\Pi_T}$. We can define a B{\"u}chi automaton as
\mbox{${B:= (Q_B, q_0, \Pi_B, \delta_B, Q_f)}$}, 
where (i) $Q_B$ is a finite set of states, (ii) $q_0 \in Q_B$ is the initial state, (iii) $\Pi_B = 2^{\Pi_T}$ is the set of input symbols accepted by the automaton, (iv) $\delta_B \subseteq Q_B \times \Pi_B \times Q_B$ is a transition relation, and (v) $Q_f \subseteq Q_B$ is a set of final states.
An accepting state in the B\"uchi automaton is the one that needs to occur infinitely often on an infinite length string consisting of symbols from $\Pi_B$ to get accepted.

\begin{example}
Figure~\ref{exampleAutomata} shows the B\"uchi automaton for an LTL task $\square(\Diamond P_{1} \land \Diamond P_{2} \land \neg P_{3})$, which means that the robots should always repeat visiting pick up locations $P_2$ and drop location $P_1$, and always avoid locations $P_3$. Here, $q_0$ is the start state as well as the final state. It informally depicts the steps to be followed in order to compete the task $\phi$. The transitions $q_1 \rightarrow q_2 \rightarrow q_0$ leads us to visit a state 
where $P_1 \land \lnot P_2 \land \lnot P_3$ is satisfied by going through only those states which satisfy $\lnot P_1 \land \lnot P_3$ and then go to state where $P_2 \land \lnot P_3 $ is satisfied using states which  satisfy $\lnot P_2 \land \lnot P_3$. This way, we can also understand the meaning of the other transitions. 
\end{example}

\subsubsection{Product Automaton}\label{productAutomaton}
\sloppy{ The product automaton $P$ between the joint transition system $T$ and the B{\"u}chi automaton $B$ is defined as
\mbox{$P := (S_P, S_{P,0}, E_P, F_P, w_p)$}, 
    where (i) $S_P = S_T \times Q_B$, (ii) $S_{P,0}:= (s_0, q_0)$ is an initial state, (iii) $E_P \subseteq S_P \times S_P$, where $((s_i,q_k) ,(s_j,q_l)) \in E_P$ if and only if $(s_i, s_j)\in E_T$ and $\left(q_k, L_T\left(s_j\right),q_l\right) \in \delta_B$, (iv) $F_P := S_T \times Q_f $ set of final states, and (v) $w_P : E_P \rightarrow \mathbb{N}_{>0}$ such that $w_P((s_i,q_k) ,(s_j,q_l)) := w_T(s_i, s_j)$}. To generate a trajectory in $T$ which satisfies LTL query, we can refer $P$.
Refer~\cite{belta5650896} for examples.

\shortversion{
\subsection{Problem Definition} 
\label{subsec-problem}

\sloppy Consider a team of robots moving in a static workspace $\mathcal{W}$ represented as transition systems $\{ T_1,...,T_n\}$ and their collective motion is modeled as a joint transition system $T$. A run over the transition system $T$ starting at initial state $s_0$ defines the trajectory of the robots in the $\mathcal{W}$.
Suppose the robots are given a task in the form of an LTL query $\phi$ over $\Pi_T$, which needs to be completed collectively by them repetitively and infinitely many times. 
We construct a B{\"u}chi automaton $B$ from $\phi$. Let $\Pi_c = \{ c \:|\: c \in \Pi_B$ and $ \exists \delta_B(q_i , c) =  q_j$ where, $q_i \in Q_B$ and $q_j \in Q_f  \}$. Let $F_{\pi} = \{ s_i \: | \: s_i \in S_T$ and $ s_i \vDash \pi_j$ where $\pi_j \in \Pi_c \}$. $F_\pi$ represents a set of all the possible final states. A final state signifies the completion of task $\phi$. Our objective is to find an infinite length path $\mathcal{R} = s_0,s_1,s_2,...$ over $T$ which satisfies $\phi$ and so, there exists $f \in F_\pi$ which occurs on $\mathcal{R}$ infinitely many times. Such path can be divided into two components namely  \emph{prefix} ($\mathcal{R}_{pre}$) and \emph{suffix} ($\mathcal{R}_{suf}$)~\cite{BaierK08}. A prefix is a finite run from the initial state $s_0$ to an accepting state $f \in F_\pi$ and a suffix is a finite length run starting and ending at $f$ reached by the prefix, and containing no other occurrence of $f$. The suffix is repeated periodically and infinitely many times to generate an infinite length run $\mathcal{R}$. Thus, we can represent run $\mathcal{R}$ as $\mathcal{R}_{pre}. \mathcal{R}_{suf}^\omega$, where $\mathcal{R}_{pre} = s_0,s_1,s_2,...,s_p$ be a prefix and $\mathcal{R}_{suf} = s_{p+1},s_{p+2},...,s_{p+r}$ be a suffix, $s_{p+r} = s_p$ and $\omega$ denotes the suffix being repeated infinitely many times.

The cost of such run can be minimized if we minimize the cost of the suffix, which can be given as 
\begin{equation} \label{eq3}
 \mathcal{C}(\mathcal{R})= \mathcal{C}(\mathcal{R}_{suf}) = \sum_{i=p+1}^{p+r-1} w_T(s_i, s_{i+1})
\end{equation}

\begin{problem}
Given a joint transition system $T$ capturing the motion of the team of robots in workspace $\mathcal{W}$ and an LTL formula $\phi$ representing the task given to the robots, find an infinite length run $\mathcal{R}$ in prefix-suffix form over $T$ which minimizes the cost function~\eqref{eq3}.
\end{problem}

\noindent
\textbf{Note:} If we want to deal with the LTL specifications that can be satisfied by the finite trajectories (for example, $\diamondsuit a$ or $a_1 U a_2$), we can define a cost function that captures the cost of the prefix, denoted by $\mathcal{C}(\mathcal{R}_{pre})$.
}

\longversion{
\subsection{Problem Definition}
\sloppy Consider a team of robots represented as transition systems $\{ T_1,...,T_n\}$, moving in a static workspace $\mathcal{W}$ and their collective motion is modeled as a joint transition system $T$. A run over the transition system $T$ starting at initial state $s_0$ defines the trajectory of the robots in the $\mathcal{W}$.
Suppose, the robots are given a task in the form of an LTL query $\phi$ over $\Pi_T$ which needs to be completed collectively by them repetitively and infinitely many times. 
We construct a B{\"u}chi automaton $B$ from $\phi$. Let $\Pi_c = \{ c \:|\: c \in \Pi_B$ and $ \exists \delta_B(q_i , c) =  q_j$ where, $q_i \in Q_B$ and $q_j \in Q_f  \}$. Let $F_{\pi} = \{ s_i \: | \: s_i \in S_T$ and $ s_i \vDash \pi_j$ where $\pi_j \in \Pi_c \}$. $F_\pi$ represents a set of all the possible final states (final state is the last state in  the complete trajectory in $T$ which satisfies $\phi$) to be visited by the team on the path to complete the task. Our objective is to find the path in $T$(which represents the trajectories of the robots in $\mathcal{W}$) in the form of cycle with minimum cost and also which completes the task. Such path will always contain one of the states from $F_\pi$.

Let us assume that there exists at least one run over $T$ which satisfies $\phi$. Let $\mathcal{R} = s_0,s_1,s_2,...$ be an infinite length run/path over $T$ which satisfies $\phi$ and so there exists $f \in F_\pi$ which occurs on $\mathcal{R}$ infinitely many times. From $\mathcal{R}$, we can extract all the time instances at which $f$ occurs. Let $t_\mathcal{R}^f(i)$ denotes the time instance of $i^{th}$ occurrence of state $f$ on $\mathcal{R}$.
Our goal is to synthesize an infinite run $\mathcal{R}$ which satisfies the LTL formula $\phi$  and minimizes the cost function 
\begin{equation} \label{eqA1}
\mathcal{C}(\mathcal{R})=\limsup_{i \rightarrow+\infty} \sum_{k=t_\mathcal{R}^f(i)}^{t_\mathcal{R}^f(i+1)-1} w_T(s_k, s_{k+1})
\end{equation}

\subsubsection{Prefix-Suffix Structure} The accepting run  $\mathcal{R}$ of infinite length can be divided into two components namely  \emph{prefix} ($\mathcal{R}_{pre}$) and \emph{suffix} ($\mathcal{R}_{suf}$). A prefix is a finite run from the initial state of the robot to an accepting state $f \in F_\pi$ and a suffix is a finite length run starting and ending at $f$ reached by the prefix, and containing no other occurrence of $f$. This suffix will be repeated periodically and infinitely many times to generate an infinite length run $\mathcal{R}$. So, we can represent run $\mathcal{R}$ as $\mathcal{R}_{pre}. \mathcal{R}_{suf}^\omega$, where $\omega$ denotes the suffix being repeated infinitely many times.

\noindent \emph{Lemma 3.1: }
For every run $\mathcal{R}$ which satisfies the LTL formula $\phi$ and minimizes cost function \eqref{eqA1}, there exists a run $\mathcal{R}_c$ which satisfies $\phi$, minimizes cost function (\ref{eqA1}) and is in prefix-suffix structure. Refer~\cite{belta5650896} for the proof. 

The cost of such run $\mathcal{R}_c$ is the cost of its suffix. So, now our goal translates to determining an algorithm which finds minimum cost suffix run starting and ending at a state $f \in F_\pi$ and having a finite length prefix run starting at initial state $s_0 \in S_T$ and ending at $f$. 
So, let $\mathcal{R} = \mathcal{R}_{pre}.\mathcal{R}_{suf}^\omega$, where $\mathcal{R}_{pre} = s_0,s_1,s_2,...,s_p$ be a prefix and $\mathcal{R}_{suf} = s_{p+1},s_{p+2},...,s_{p+r}$ be a suffix, where $s_{p+r} = s_p$. 
We can redefine the cost function given in \ref{eqA1} as
\begin{equation} \label{eqA3}
 \mathcal{C}(\mathcal{R})= \mathcal{C}(\mathcal{R}_{suf}) = \sum_{i=p+1}^{p+r-1} w_T(s_i, s_{i+1})
\end{equation}

\begin{question}
Given a joint transition system $T$ capturing the motion of the team of robots in workspace $\mathcal{W}$ and an LTL formula $\phi$ representing the task given to the robots, find an infinite length run $\mathcal{R}$ in prefix-suffix form over $T$ which minimizes the cost function \eqref{eqA3}.
\end{question}

\noindent
\textbf{Note:} If we want to deal with the LTL specifications that can be satisfied by the finite trajectories (for example, $\diamondsuit a$ or $a_1 U a_2$), we can define a cost function that captures the cost of the prefix, denoted by $\mathcal{C}(\mathcal{R}_{pre})$.
}

\shortversion {
\subsection{Baseline Solution Approach}
The state-of-the-art solution to the above problem~\cite{UlusoySDBR13} uses the automata-theoretic model checking approach.  It computes the product automaton of $T$ and $B$ and then uses Dijkstra's shortest path algorithm to compute the required minimum cost suffix run having a valid prefix (see Appendix~\ref{AppendixBaselineApproach} for details). 
The size of $T$ increases exponentially with the increase in the number of robots and also the workspace size. It consumes a huge amount of memory, which becomes a hindrance to scaling up this algorithm.
In the next section, we present our algorithm \textsf{MT*} and use the algorithm proposed in~\cite{UlusoySDBR13} as the baseline for quantitative comparison. 
}

\longversion {
\subsection{Baseline Solution Approach}
The baseline solution to above problem uses the automata-theoretic model checking approach~\cite{UlusoySDBR13}, the steps of which are outlined in the Algorithm~\ref{baselineSolution}.

\begin{algorithm2e}
\SetAlgoLined
\DontPrintSemicolon
\setstretch{0.9}
\textbf{Input:} Transition systems $\{ T^1,...,T^n\}$, $\phi$: An LTL formula\;
\textbf{Output: }A set of runs $\{ \mathcal{R}^1,...,\mathcal{R}^n \}$that satisfies $\phi$\;
\BlankLine
Construct a joint transition system $T$\;
Convert $\phi$ to a B\"uchi automaton $B$\;
Compute the product automaton $P = T \times B$\;
\For{all $f \in F_P$ }{
$\mathcal{R}_f^{suf} \gets \mathtt{Dijkstra's\_Algorithm}(\: \emph{f} \:,\:  \emph{f}\:)$\;
$\mathcal{R}_f^{pre} \gets \mathtt{Dijkstra's\_Algorithm}(\: S_{P,0} \:,\:  \emph{f}\:)$
}
$\mathcal{R}_P^{suf} \gets$ minimum of all $\mathcal{R}_f^{suf}$ \;
$\mathcal{R}_P^{pre} \gets$ prefix of $\mathcal{R}_P^{suf}$ \;
$\mathcal{R}_P = \mathcal{R}_P^{suf} . \mathcal{R}_P^{pre} $\;
Project $\mathcal{R}_P$ over $T$ to compute $\mathcal{R}_T$\;
Project $\mathcal{R}_T$ over $\{ T^1,...,T^n\}$ to obtain runs $\{\mathcal{R}^1,...,\mathcal{R}^n\}$
\caption{Baseline\_Solution}
\label{baselineSolution}
\end{algorithm2e} 

The first step in this algorithm is to compute
the joint transition system $T$ from the transition systems of the individual robots $T^i$. Then we compute the B\"uchi automaton from the given LTL query $\phi$. We then compute the product automaton of $T$ and $B$. In this product automaton, for each final state $f \in F_P$, we find a prefix run starting from initial state $S_{P,0}$ to $f \in F_P$ and then find minimum cost cycle starting and ending at $f$ using Dijkstra's algorithm. We then choose the prefix-suffix pair with the smallest $\mathcal{C}(\mathcal{R}_P)$ cost i.e. the pair with smallest suffix cost, and project it on $T$ to obtain the run $\mathcal{R}_T$ which represents the joint motion all the robots in $\mathcal{W}$. We then  project $\mathcal{R}_T$ over individual $T^i$ to obtain the  of the individual robots $\mathcal{R}^i$. The trajectory $\mathcal{R}^i$ for the $i$-th robot provides us with the cyclic trajectory which the robot can follow repetitively to complete the given task $\phi$ repetitively.  

In the following section, we present MT$^*$ algorithm that provides a significantly improved running time for generating an optimal trajectory satisfying a given LTL query.
}

\section{\textsf{MT*} Algorithm}

In \textsf{MT*}, we divide a complex LTL path planning problem into simpler problems systematically, which can be solved individually and then combined to solve the original problem optimally. 
\textsf{MT*} only computes a reduced version of the product graph $P$, which we call the \emph{Abstract Reduced Graph} $G_r$. Its size is significantly smaller compared to $P$ and thus is faster and consumes less memory.


\begin{algorithm2e}
\setstretch{1.03}
\DontPrintSemicolon
\textbf{Input:} Transition systems $\{  T^1,...,T^n\}$, $\phi$: An LTL formula\;
\textbf{Output: }A run $\langle \mathcal{R}^1,...,\mathcal{R}^n \rangle$ that satisfies $\phi$\;
\BlankLine
\BlankLine
$B (Q_B, q_{0}, \Pi_B, \delta_B, Q_f) \gets \mathtt{ltl\_to\_Buchi}$
($\phi$)\;
\For{ all $q_i,q_j \in Q_B$, where $\delta_B(q_i,c_{pos}) = q_j$ }
{
    $S_*[c_{pos}]$ =  $\mathtt{Abstract\_Distant\_Neighbours}(c_{pos})$   
}
$G_r(S_r, v_{0}, E_r, F_r,\mathcal{N})\gets \mathtt{Generate\_Redc\_Graph(B,T)}$\label{T* line 2}\;
\For{all $f \in F_r$}{
    \For{each simple cycle $C_f$ containing $f$ in $G_r$ }
    {
        $B' \gets \mathtt{Extract\_Buchi\_Trans\_From\_C_f}(C_f,B)$\;
        \For{$i \gets \{ 1,..,n\}$}
        {
            $c^i_f \gets \mathtt{Project\_C_f\_Over\_T^i}(C_f, i, G_r)$\;
            $B^i \gets \mathtt{Project\_C^i_f\_Over\_B'}(c^i_f, B', G_r)$\;
            $\mathcal{R}^i_f \gets \mathtt{Optimal\_Run}(B^i,c^i_c, T^i)$
        }
        $\mathcal{R}_f^{suf}(\langle \mathcal{R}^1_{s},..,\mathcal{R}^n_{s} \rangle) \gets \mathtt{Sync}(\langle \mathcal{R}^1_f,..,\mathcal{R}^n_f \rangle)$\;
        $\mathcal{R}_f^{pre}(\langle\mathcal{R}^1_p,..,\mathcal{R}^n_p\rangle) \gets \mathtt{Compt\_Prefix}(f, B, G_r)$\;
    }
}
$\mathcal{R}_P^{suf} \gets \argmin\limits_{\mathcal{R}_f^{suf} \text{ with a valid prefix}}  \mathcal{C}\left(\mathcal{R}_f^{suf}\right)$\;
$\mathcal{R}_P^{pre} \gets$ prefix of $\mathcal{R}_P^{suf}$ \;
$\mathcal{R}_P = \mathcal{R}_P^{pre} . \mathcal{R}_P^{suf} $\;
Project $\mathcal{R}_P$ over $T$ to compute $\mathcal{R}_T$\;
Project $\mathcal{R}_T$ over $\{  T^1,...,T^n\}$ to obtain runs $\langle \mathcal{R}^1,...,\mathcal{R}^n\rangle$
\BlankLine
\BlankLine
\SetKwProg{myproc}{Procedure}{}{}\label{p1}
\myproc{ \textbf{Generate\_Redc\_Graph}($B, T$) } {

$v_{init} \gets v_0(s_0, q_0)$\;
    let $Q$ be a queue data-structure\;
    Initialize empty reduced graph $G_r$\;
    label $v_{init}$ as discovered and add it to $S_r$\;
    
    $Q.\mathtt{enqueue}(v_{init})$\;
    \While{$Q$ is not empty}
    {
        $v_i(s_i,q_i) \gets Q.\mathtt{dequeue}( )$\;
    
        \If{$\exists \delta_B(q_i,c_{neg})=q_i \text{ and } \nexists (\delta_B(q_i, c_{neg}) = q_j \text{ such that } q_i \neq q_j )$ }
        {
            \For{all $v_l(s_l,q_l)$ such that $s_l \in S_*[c_{pos}]$ and $\delta_B(q_i, c_{pos}) = q_l$ }
            {
                $E_r \gets (v_i, v_l)$\;
                $\mathcal{N}(v_i,v_l) \gets false$\;
                \If{$v_l$ is not labelled as discovered}
                {
                    label $v_l$ as $discovered$, add it to $S_r$\; 
                    $Q.\mathtt{enqueue}(v_l)$ 
                }
            }
        }
        \Else{
            \For{all $v_l(s_l,q_l)$ such that $s_l \in S_*[c_{pos}]$ and $\delta_B(q_i, c_{pos}) = q_l$ }
            {
                $E_r \gets (v_i, v_l)$\; $\mathcal{N}(v_i,v_l) \gets true$\;
                \If{$v_l$ is not labelled as discovered}
                {
                    label $v_l$ as $discovered$, add it to $S_r$\; 
                    $Q.\mathtt{enqueue}(v_l)$ 
                }
            }
        
            \For{all $q_l$ such that $\exists \delta_B(q_i, c_{neg})=q_l$ }
            {
                $v_l \gets (s_*,q_l)$\;
                $E_r \gets (v_i, v_l)$\; $\mathcal{N}(v_i,v_l) \gets true$\;
                \If{$v_l$ is not labelled as discovered}
                {
                    label $v_l$ as $discovered$, add it to $S_r$\; 
                    $Q.\mathtt{enqueue}(v_l)$ 
                }
                
            }
        }
      
    }
    return $G_r$
}
\caption{\textsf{MT*}}
\label{mtStar}
\end{algorithm2e}

\subsection{Abstract Reduced Graph} \label{ARG}
We explain the intuition behind the construction of abstract reduced graph $G_r$ using a single robot example. 
\begin{example}
Consider a robot moving in workspace $\mathcal{W}$ shown in Figure~\ref{exampleGrid} and has been given an LTL task $\square(\Diamond P_{1} \land \Diamond P_{2} \land \lnot P_{3})$ whose B\"uchi automaton $B$ is shown in Figure~\ref{exampleAutomata}. Let $T^1$ be the transition system of the robot constructed from $\mathcal{W}$. For one robot system, the joint transition system $T$ will be same as $T^1$.
Consider a product automaton $P$ of $T$ and $B$. Suppose $s_0 = \langle (4,7)\rangle $ and therefore $S_{P,0} = (\langle (4,7)\rangle , q_0)$. Now, from here, we must use the transitions in the B\"uchi automaton to find the path in $T$ in the prefix-suffix form. Suppose we find such a path on which we move to state $(\langle (4,6)\rangle , q_1)$ from $(\langle (4,7)\rangle , q_0)$ as per the definition of the product automaton. From $(\langle (4,6)\rangle , q_1)$,
we must visit a location where $P_1 \land \lnot P_2 \land \lnot P_3$ is satisfied so that we can move to B\"uchi state $q_2$ from $q_1$. All the intermediate states till we reach such a state must satisfy $\lnot P_1 \land \lnot P_3$ formula. Suppose we next move from $(\langle (4,6)\rangle ,q_1)$ to $(\langle (6,6)\rangle , q_2)$  on $P$ which satisfies $P_1 \land \lnot P_2 \land \lnot P_3$ and this path is $(\langle (4,6)\rangle , q_1)$ $\rightarrow$ $(\langle (4,5)\rangle , q_1)$ $\rightarrow$ $(\langle (4,4)\rangle , q_1)$ $\rightarrow$ $...$ $\rightarrow$ $(\langle (6,5)\rangle , q_1)$ $\rightarrow$ $(\langle (6,6)\rangle , q_2)$. 
On the path from $(\langle (4,6)\rangle , q_1)$ to $(\langle (6,6)\rangle , q_2)$, all the intermediate nodes satisfy the self-loop transition condition on $q_1$.
We can consider the self-loop transition condition $\lnot P_1 \land \lnot P_3$ over $q_1$ as the constraint which must be satisfied by the intermediate states while completing a task of moving to the location satisfying the transition condition from $q_1$ to $q_2$. 
Using this as an abstraction method over the product automaton, we directly add an edge from state $(\langle (4,6)\rangle ,q_1)$ to state $(\langle (6,6)\rangle ,q_2)$ in the reduced graph assuming that there exists a path between these two states. We explore this path opportunistically only when it is required. This is the first idea behind \textsf{MT*} algorithm.
\end{example}

Throughout this paper, we call an atomic proposition with negation a \emph{negative proposition} and an atomic proposition without negation a \emph{positive proposition}. For example, $\lnot P_2$ is a negative proposition and $P_2$ is a positive proposition. We divide the transition conditions in $B$ into two types. A transition condition which is a conjunction of all negative propositions is called a \emph{negative transition condition} and is denoted by $c_{neg}$. The one which is not negative is called a \emph{positive transition condition}, and is denoted by $c_{pos}$. For example, $\lnot P_1 \land \lnot P_3 $ is a negative whereas $P_1 \land \lnot P_2 \land \lnot P_3$ is a positive transition condition. 
For any transition, we can consider all the positive propositions as the task to be completed by all the robots collectively and all the negative transitions as constraints that must be followed by all the robots. For example,  in $P_1 \land P_2 \land \lnot P_3$, some robot must visit a location where $P_1$ is satisfied. At the same time, some robot must visit a location where $P_2$ is satisfied, and all the robots must satisfy $\lnot P_3$ to satisfy transition $P_1 \land P_2 \land \lnot P_3$ completely and collectively. Let us move to the second idea. 
\begin{example}
Consider that $c_{pos} = P_2 \land \lnot P_3$ be a transition condition from $B$. Suppose here we are planning the paths for two robots. So, to satisfy this transition, one of the robots must go to a location where $P_2 \land \lnot P_3$ is satisfied.  There are two ways to achieve it. The first one is when robot 1 reaches a location on $\mathcal{W}$ where $P_2 \land \lnot P_3$ is satisfied, and robot 2 is at the location where $\lnot P_3$ is satisfied. Then we can say that $P_2 \land \lnot P_3$ satisfied by both the robots. And second one is when robot 2 satisfies $P_2 \land \lnot P_3$ and robot 1 satisfies $\lnot P_3$. In the first case, robot $1$ could be at either $(0,7)$ or $(7,0)$ which satisfies $P_2 \land \lnot P_3$ and robot two could be at any of 52 locations which are obstacle-free and satisfy $\lnot P_3$. The number of ways in which two robots can satisfy the first case is $52*2 = 104$. Similarly, in another $104$ ways, the robots can satisfy the second case. 
We represent all these $104+104=208$ satisfying configurations symbolically as 
\scalebox{0.9}{$\{\langle(0,7),(*,*)\rangle , \langle(7,0),(*,*)\rangle, \langle(*,*),(0,7)\rangle, \langle(*,*),(7,0)\rangle\}$}. Here, $((0,7),(*,*))$ says that robot 1 is at location $(7,0)$ whereas robot 2 could be at any obstacle-free location which satisfies $\lnot P_3$. We call this set as abstract neighbour set for transition condition $c_{pos}$ and represent it as $S_*[c_{pos}]$. For a single robot, we represent an unknown state symbolically as $s_*^i := (*,*)$, where $i \in \{ 1,...,n\}$. If the locations of all the robots are unknown, then we represent such state as $s_* := \langle s_*^1,...,s_*^n \rangle $. We can easily compute $S_*[c_{pos}]$ for any $c_{pos}$ transition condition by distributing all the task propositions among the robots in all possible ways and denoting the locations for robots who do not receive any task from $c_{pos}$ as $(*,*)$. While computing $S_*$, we also store the tasks and the constraints that each robot satisfies in a particular configuration using a map $L_*$. For example,  $S_*[c_{pos} = P_2 \land \lnot P_3]$ = \scalebox{0.9}{$\{  \langle(0,7),(*,*)\rangle , \langle(7,0),(*,*)\rangle, \langle(*,*),(0,7)\rangle, \langle(*,*),(7,0)\rangle  \}$} and $L_*[((0,7),(*,*))] = \{  \{ P_2 , \lnot P_3  \}, \{ \lnot P_3\} \}$ and so on.
\end{example}
Using the above ideas, we can represent the product graph symbolically as a significantly smaller abstract reduced graph. While constructing an abstract reduced graph, we add an edge from node $v_i(s_i, q_i)$ to some node using the following rules:\\
\noindent 
\textbf{$Condition:$ $\exists \delta_B(q_i,c_{neg})=q_i \text{ and } \nexists \delta_B(q_i, c_{neg}) = q_j , q_i \neq q_j $}, $i.e.$, if there exists a negative self loop over $q_i$ and there does not exist any negative transition from $q_i$ to some other state in the B\"uchi automaton.

\begin{figure}[t]
      \centering
      \includegraphics[scale=0.11]{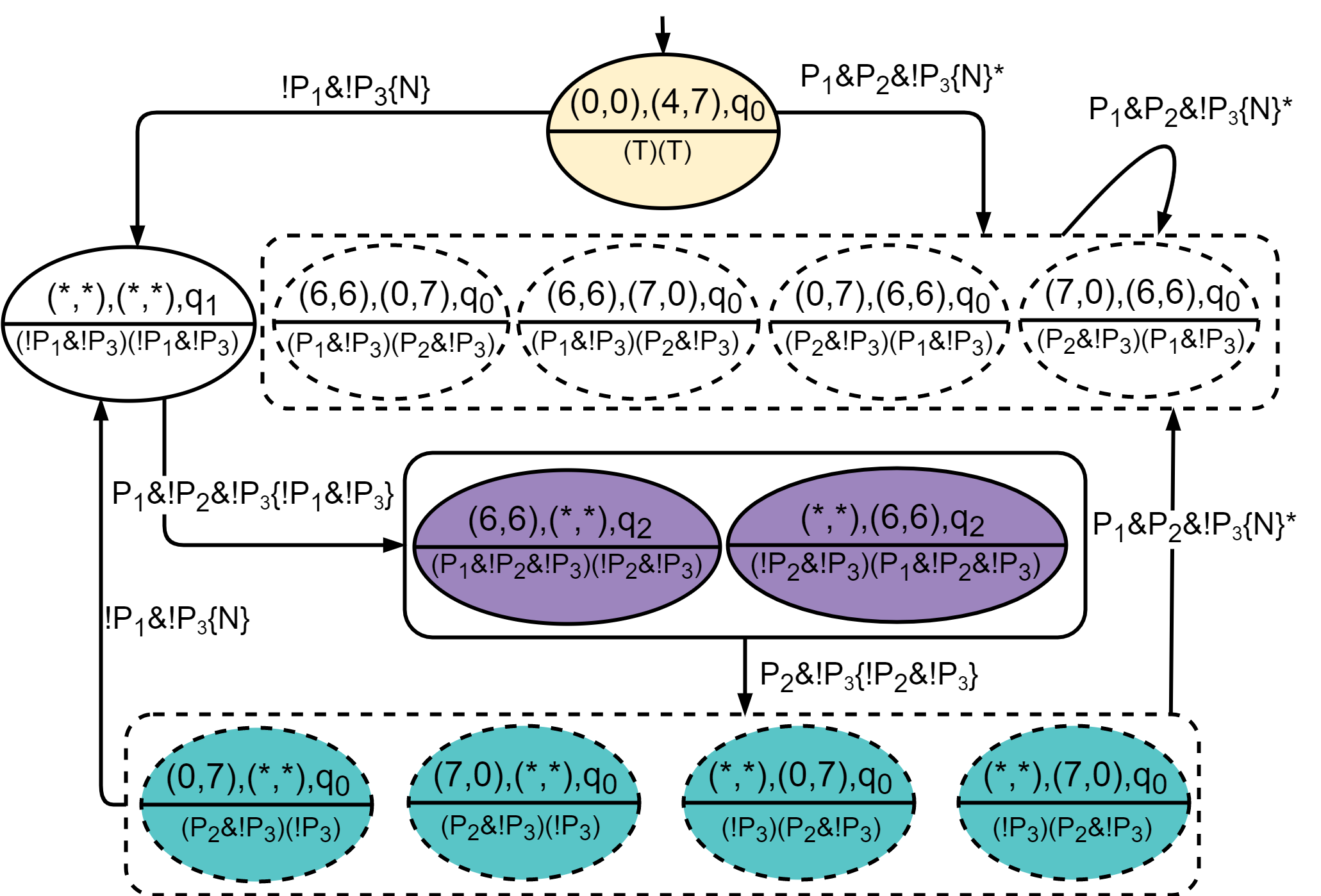}
      \caption{Abstract Reduced Graph for a $2$ Robot System Having Workspace and B\"uchi Automaton from Figure \ref{ws}}
      \label{AbstRedGraph}
\end{figure}

\textbf{\emph{If Condition is true then}} add an edge from $v_i(s_i,q_i)$ to all $v_l(s_l,q_l)$ such that $\exists \delta_B(q_i, c_{pos}) = q_l$ and $s_l \in S_*[c_{pos}]$. Here, $q_i$ and $q_l$ can be the same. In short, in this condition we add all the nodes as neighbours to $v_i$ which satisfy an outgoing $c_{pos}$ transition from $q_i$ and skip nodes which satisfy  $c_{neg}$ self loop transition assuming that $c_{neg}$ self loop transition can be used to find the actual path from $v_i$ to $v_l$ later in the algorithm. We use $\mathcal{N}$ to keep track neighbour information in $G_r$. $\mathcal{N}(v_a,v_b) = true$ says that $v_b$ must be a neighbour of $v_a$ in $T$. In this condition, we set $\mathcal{N}(v_i,v_l)$ to $false$. 
    It says that $v_i$ and $v_l$ may not be actual neighbours in the actual product graph $P$. For example, consider an edge between $v_i =(\langle(6,6)(*,*)\rangle, q_2)$ and $v_l =(\langle(7,0)(*,*)\rangle, q_0)$ in Figure \ref{AbstRedGraph}. Here, we say that the path between $(6,6)$ and $(7,0)$ could be established through $c_{neg}$ transition condition.
    Similarly, whatever value we fill for unknowns $(*,*)$, intermediate path between the two must satisfy $c_{neg}$ condition.
    We call this way of adding neighbours as \emph{distant neighbour way/condition}.
    
\textbf{\emph{If Condition is  false then}} the same as above, we add an edge from $v_i$ to all $v_l(s_l,q_l)$ such that $\exists \delta_B(q_i, c_{pos}) = q_l$ and $s_l \in S_*[c_{pos}]$. And for all $c_{neg}$ outgoing transitions from $q_i$ to some $q_l$, we add an edge from $v_i$ to  $v_m(s_*,q_l)$. Here $s_*:=\langle s_*^1,...,s_*^n \rangle$ where, $s_*^i = (*,*)$. Here, $q_i$ and $q_l$ can be the same. We call this way of adding neighbours as \emph{product automaton way/condition}. For all these transitions, we set $\mathcal{N}(v_i, v_l)$ to  $true$ which says that $v_i$ and $v_l$ are actual neighbours in $T$. At this moment we do not know the complete value of the nodes in $v_i$ and $v_l$. For example, as shown in Figure \ref{AbstRedGraph}, $v_i$ could be$(\langle (*,*)(7,0)\rangle ,q_0)$ and $v_l$ could be $(\langle (6,6)(7,0)\rangle ,q_0)$. Here for robot 2, $(7,0)$ from $v_i$  is the neighbour of $(7,0)$ form $v_l$. But, robot 1's location in $v_i$ is not known (represented as $(*,*)$) and is known in $v_l$ as $(6,6)$. So, later in the algorithm, whenever we fill this unknown value, we have to ensure that it is neighbour of $(6,6)$ in $T^1$. 
We formally define the \emph{Abstract Reduced Graph} for the transition system $T$ and the B\"uchi automaton $B$  as 
\mbox{$G_r := (S_r, v_{init}, E_r, F_r)$},
where (i) $S_r$ the set of vertices added as per the above rules, (ii) $v_{init}= (s_0, q_0)$ is an initial state, (iii) $E_r \subseteq S_r \times S_r$, is a set of edges added as per the above conditions, (iv) the set of final states $F_r \subseteq S_r \text{ and } v_i(s_i,q_i) \in F_r \text{ iff } q_i \in Q_B$, and (v) $\mathcal{N}$ stores the neighbour information.

In procedure $\mathtt{Generate\_Redc\_Graph}$ of Algorithm~\ref{mtStar}, we run the Breadth-First-Search (BFS) algorithm starting from node $(s_0, q_0)$ and add the neighbours using above rules. 

\begin{example}
The abstract reduced graph generated for a two robot system over a workspace $\mathcal{W}$ and B\"uchi automaton $B$ from Figure~\ref{ws} is shown in Figure \ref{AbstRedGraph}. All the nodes enclosed in a rectangle represent an abstract neighbour set $S_*$ for some transition. Edges have transition conditions written on them. ${N}$ mentioned on the transition represents $\mathcal{N}$ value for that transition.
$^*$ on the transition says that this transition is applicable only if the nodes are actual neighbours. 

Consider an edge from $v_1$ $=$ $(\langle (0,0),(4,7)\rangle ,q_0)$ to $v_2$ $=$ $(\langle (6,6),(0,7)\rangle ,q_0)$ with transition condition $P_1 \land P_2 \land \lnot P_3\{N\}^*$. Here $\{N\}$ says that $v_1$ and $v_2$ should be neighbours but as these two are not neighbours of each other in $T$, an edge cannot exist between $v_1$ and $v_2$. The symbol $^*$ here says that the edge from $v_1$ $=$ $(\langle (0,0),(4,7)\rangle ,q_0)$ to the other nodes is valid if the other nodes are neighbours of $v_1$. In Figure \ref{AbstRedGraph}, the transition condition accompanied with $c_{neg}$ condition in curly braces $\{ \}$ represents a $c_{neg}$ transition that must be used to reach next node. In every node, below the robot coordinates, we show the $L_*$ value for that node.

We explain the construction of the Abstract Reduced Graph shown in Figure \ref{AbstRedGraph}. Initially, robot 1 is at location $(0,0)$ and robot 2 is at $(4,7)$. So, here $s_0$ $=$ $\langle s_0^1, s_0^2\rangle$ $=$ $\langle (0,0),(4,7)\rangle $.
We start BFS  with node $v_0$ $=$ $(s_0, q_0)$ $=$ $(\langle (0,0),(4,7)\rangle ,q_0)$. For the first time $v_i$ $=$ $v_0$ is de-queued from the queue $Q$, so we add the neighbours of $v_0$ to $G_r$. 
Here, $q_0$ does not have $c_{neg}$ type self transition loop in B. So, we first add all $c_{pos}$ satisfying nodes as neighbours of $v_0$. Here, there exists a transition from $q_0$ to $q_0$ on $c_{pos} = P_1 \land P_2 \land \lnot P_3$. We add edges from $v_0$ to all the nodes in $S_*[(P_1 \land P_2 \land \lnot P_3)]$ which are $\langle (6,6)(0,7)\rangle , \langle (6,6)(7,0)\rangle , \langle (0,7)(6,6)\rangle $, and $\langle (7,0)(6,6)\rangle $ with $q_0$ as the B\"uchi state. As these nodes have been added as per product automaton condition, these nodes must be neighbours to $v_0$. However, None of these nodes are neighbours to $v_0$. So none of these edges will be actually added to $G_r$. We have only shown these transitions in Figure \ref{AbstRedGraph} for the sake of completeness and understanding of the readers. This kind of uncertainty we represent using $*$ on the transition condition. Now, there also exists a transition from $q_0$ to $q_1$ with transition condition $\lnot P_1 \land \lnot P_3$. And as this is a $c_{neg}$ type transition, we add an edge from $v_0$ to $(\langle (*,*),(*,*)\rangle ,q_1)$. Again this node has been added as per product automaton condition, so whatever value we choose to put in place of $\langle (*,*)(*,*)\rangle $ must be neighbour of $v_0$.
Now, suppose $v_i= (\langle (*,*),(*,*)\rangle ,q_1)$ has been de-queued from the queue $Q$. Here, there exists a negative self loop over $q_1$ with transition condition $c_{neg} = \lnot P_1 \land \lnot P_3$ and there does not exist any $c_{neg}$ transition from $q_1$ to any other state in $B$. So, we add all the nodes as neighbours to $v_i$ which satisfy $c_{pos}$ transition conditions and ignore $c_{neg}$ self loop. There exists a positive transition from $q_1$ to $q_2$ with transition condition $c_{pos} = P_1 \land \lnot P_2 \land \lnot P_3$. 
So, we add edges from $v_i$ to all the nodes in $S_*[(P_1 \land \lnot P_2 \land \lnot P_3)]$ which are $\langle (6,6),(*,*)\rangle  , \langle (*,*),(6,6)\rangle $ with B\"uchi state $q_2$. As these nodes have been added as neighbours to $v_i$ using distant neighbour condition, they may not be actual neighbours of $v_i$ in $P$. We explain this in depth in the next section. Like this, we add nodes to $G_r$. Completed $G_r$ is shown in Figure~\ref{AbstRedGraph}. 
\end{example}


\subsection{\textsf{MT*} Procedure}
\begin{figure}
      \centering
      \includegraphics[scale=0.12]{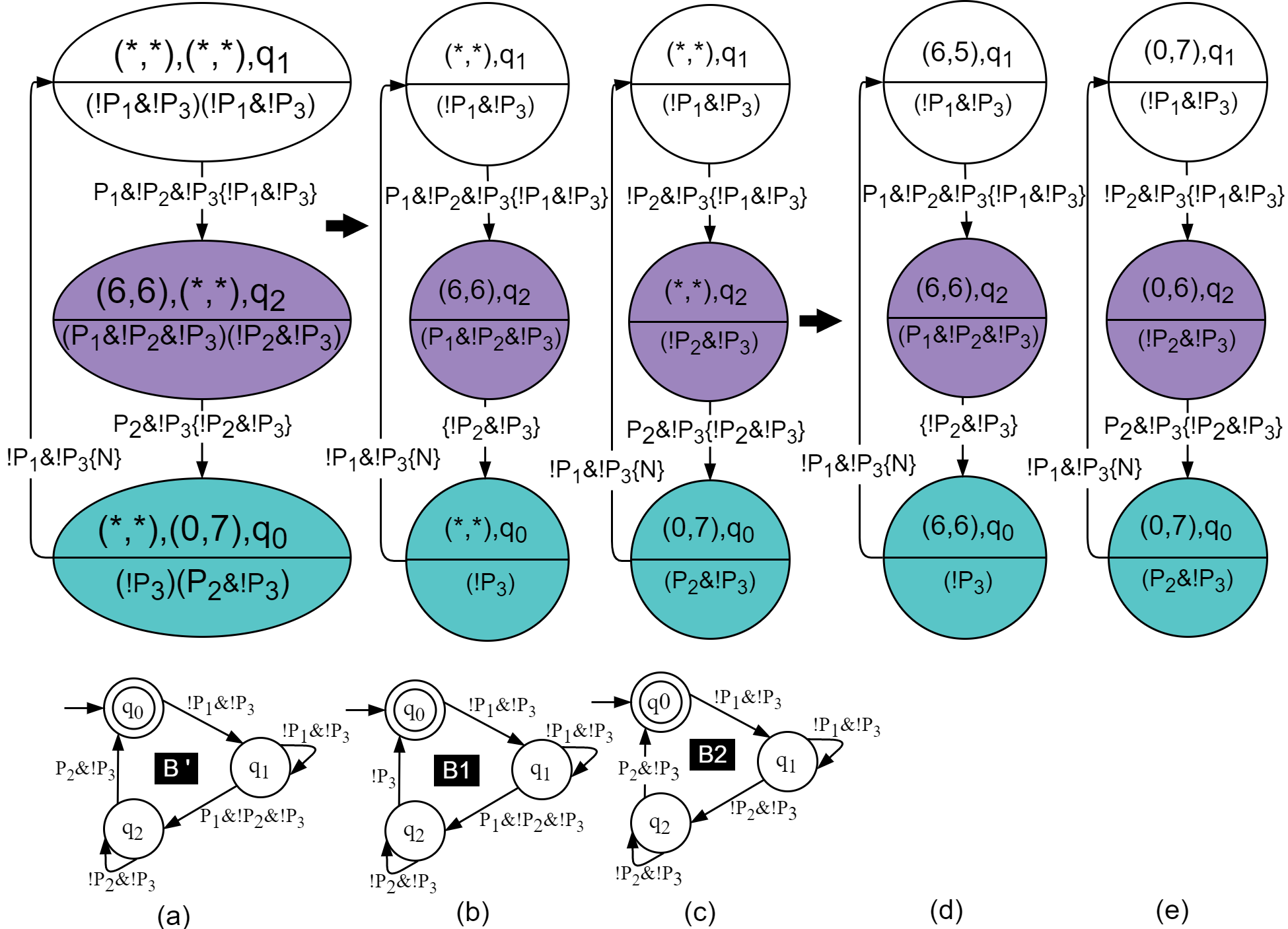}
      \caption{Suffix Computation in \textsf{MT*} Algorithm}
      \label{exampleshort}
\end{figure}
We outline all the block steps of \textsf{MT*} in Algorithm~\ref{mtStar} and explain the concepts for the major steps. 
Given the transition systems for $n$ robots $\{ T^1,...,T^n \}$ and an LTL query $\phi$, the goal is to compute a minimum cost run $\mathcal{R}^i$ over $T^i$ for $n$ robots, satisfying $\phi$ in the form of prefix and suffix.

(1) In this algorithm, we first compute the B\"uchi automaton from the given LTL task $\phi$. 
(2) Then for all $c_{pos}$ transition conditions present in $B$, we compute the abstract neighbour set $S_*[c_{pos}]$ using the procedure $\mathtt{Abstract\_Distant\_Neighbour}$() as explained in section (\ref{ARG}).
(3) Then we generate Abstract Reduced Graph ($G_r$) using procedure $\mathtt{Generate\_Redc\_Graph}$() from Algorithm \ref{mtStar}. 
(4) Final state $f_p \in F_P$ corresponds to an incoming transition to a final state in the B\"uchi automaton. In the product graph, the final state corresponds to a state where the task completes and starts again. As our task is to find a minimum cost cyclic trajectory that satisfies LTL task $\phi$, it will always contain a final state, and as the trajectory is cyclic,  if we start from $f_p$, we will again come back to it over the cyclic trajectory. $F_r$ is a symbolic representation of $F_P$. So, for each state $f \in F_r$, we compute all the possible cycles starting and ending at $f$. We choose the one with the minimum cost and reachable from $v_{init}$ as our final solution. 
(5) \emph{Optimization}: If a node other than $f$ repeats on the suffix cycle, then there exists another cycle on the suffix. We can always obtain a suffix with a smaller cost by removing this extra cycle. So, we only search for a simple cycle (in which no node is repeated except the starting and the ending node $f$). We can easily find such cycles using Depth-First Search (DFS) algorithm starting with node $f$.
(6) For each $f \in F_r$ and for each simple cycle $C_f$ starting and ending at $f$ in $G_r$, we follow the following steps. We will also use the example mentioned in Figure~\ref{exampleshort} for a better understanding of the readers.
(i) $\mathtt{Extract\_Buchi\_Trans\_From\_c_f}$( ): Each cycle $C_f$ in $G_r$ also represents a cycle of transitions in the B\"uchi automaton. For example, consider a cycle $C_f$ starting at ending at $f = (\langle (*,*),(0,7)\rangle ,q_0)$ in Figure \ref{exampleshort}(a). From this $C_f$, we can extract an automaton $B'$ which is a sub-graph of $B$ and also shown in Figure \ref{exampleshort}(a). 
(ii) After this, we decouple the joint suffix cycle into $n$ cycles individual to each robot. 
(iii) $\mathtt{Project\_C_f\_Over\_T^i}$( ): We use this procedure to project/extract $C_f$ over $T^i$ to compute the trajectory $c_f^i$ for robot $i$. In Figure \ref{exampleshort}, We decouple joint trajectory shown in sub-figure (a) into two trajectories $c_f^1$ and $c_f^2$ shown in sub-figure (b) and (c) respectively. Transitions are also divided as per the constraints allocated to the individual robots.
(iv) $\mathtt{Project\_C_f^i\_Over\_B'}$( ): Using this procedure, we derive $n$ automata $\{ B^1,...,B^n \}$  from the transitions of cycles $\{ c_f^1,...,c_f^n \}$. These automata represent the path/constraints that each robot must follow in this particular joint trajectory/task $C_f$. Automata $B^1$ and $B^2$ are shown in Figure \ref{exampleshort} (b) and (c). 
(v) $\mathtt{Optimal\_Run}$( ): In this procedure, we complete the individual incomplete trajectories $c^i_f$ for all the robots using their individual automaton $B^i$. For example, the incomplete trajectory shown in sub-figure (b) is completed using automaton $B^1$ to obtain the completed trajectory shown in sub-figure (d). In trajectory (b), we first find the first known node which is $s_{source}^1=(\langle (6,6)\rangle ,q_0)$. Then we find the next known node on the cyclic trajectory which is also $s_{dest}^1=(\langle (6,6)\rangle ,q_0)$. Then we attempt to find the path from $s_{source}^1$ to $s_{dest}^1$ with automaton $B^1$ in $T^i$ using single robot LTL pathfinding algorithms \textsf{T*}~\cite{KhalidiGS20} or $\mathtt{Optimal\_Run}$~\cite{belta5650896}. In these algorithms, as we now have a concrete goal node, we can use A$^*$ instead of Dijkstra's algorithm to improve the computation time. 
Here, there was only one known node in the trajectory. However, in general, we continue like this till we find all the unknown sub-trajectories in the suffix cycle.  
(vi) This way, the problem of finding the joint trajectory for $n$ robots has been reduced to finding $n$ trajectories for a single robot system. As the size of the single robot transition system is much smaller than the joint transition system, \textsf{MT*} produces results much faster than the state of art algorithm~\cite{UlusoySDBR13}.  
(vii) $\mathtt{Sync}$( ): When we generate the robot trajectories independently for individual robots, the generated trajectories may not be in sync. The computed individual robot trajectories can be of different lengths, and because of this, the sequence in which particular locations are visited may change and may not satisfy $\phi$. 
However, if the robots can be stopped at some location during their operation, it is straightforward to achieve synchronization.
There are two types of transitions in $G_r$. One is added using the product automaton condition, in which both the nodes should be neighbours. In this case, all the generated individual trajectories will have a consistent B\"uchi state. 
The second one is the transition added using the distant neighbour condition, in which we assume that the added node will be reached using $c_{neg}$ type self-loop. In such cases, generated individual trajectories could be of different lengths. 
\begin{example}
Consider an example of some hypothetical LTL query in which there is an edge from $(\langle (6,6)(6,6)\rangle ,q_1)$ $\xrightarrow[]{ P_2 \land \lnot P_3 }$ $(\langle (7,0),(0,7)\rangle ,q_2)$ in $G_r$ with $c_{neg} = \lnot P_2 \land \lnot P_3$ type negative self loop over $q_1$. In this scenario, individual trajectories generated using procedure $\mathtt{Optimal\_Run()}$ from $(\langle (6,6)\rangle ,q_1)$ to $(\langle (7,0)\rangle ,q_2)$ for robot 1 is $ \rho_1$ $=$ $(\langle (6,6)\rangle ,q_1)$ $\xrightarrow[]{\lnot P_2 \land \lnot P_3 }$ $(\langle (7,6)\rangle ,q_1)$ $\xrightarrow[]{ \lnot P_2 \land \lnot P_3}$ $...(\text{4 nodes})...$ $\xrightarrow[]{\lnot P_2 \land \lnot P_3 }$ $(\langle (7,1)\rangle ,q_1)$ $\xrightarrow[]{ P_2 \land \lnot P_3 }$ $(\langle (7,0)\rangle ,q_2)$ consisting of total 6 transitions with $c_{neg}$ transition condition and one $c_{pos}$ transition ($P_2 \land \lnot P_3$). Whereas, for robot 2, individual trajectory is $\rho_2$ $=$ $(\langle (6,6)\rangle ,q_1)$ $\xrightarrow[]{\lnot P_2 \land \lnot P_3 }(\langle (6,5)\rangle ,q_1)\xrightarrow[]{ \lnot P_2 \land \lnot P_3 }$ $...(\text{14 nodes})...$ $\xrightarrow[]{\lnot P_2 \land \lnot P_3 }$ $(\langle (1,7)\rangle ,q_1)$ $\xrightarrow[]{ P_2 \land \lnot P_3 }$ $(\langle (0,7)\rangle ,q_2)$ with 16 $c_{neg}$ transitions and 1 $c_{pos}$ transition ($P_2 \land \lnot P_3$). While synchronizing these two trajectories, robot 1 waits at $(\langle (6,6)\rangle ,q_1)$ in $\rho_1$ (shorter trajectory) to extend it as $\rho_1'$ $=$ $(\langle (6,6)\rangle ,q_1)$ $\xrightarrow[]{\lnot P_2 \land \lnot P_3 }$ $(\langle (6,6)\rangle ,q_1)$ $\xrightarrow[]{\lnot P_2 \land \lnot P_3 }$ $...(\text{9 nodes})...$ $\xrightarrow[]{\lnot P_2 \land \lnot P_3 }$ $(\langle (6,6)\rangle ,q_1)$ $\xrightarrow[]{\lnot P_2 \land \lnot P_3 }$ $(\langle (7,6)\rangle ,q_1)$ $\xrightarrow[]{ \lnot P_2 \land \lnot P_3 }$ $...$ $(\text{4 nodes})$ $...$ $\xrightarrow[]{\lnot P_2 \land \lnot P_3 }$ $(\langle (7,1)\rangle ,q_1)$ $\xrightarrow[]{ P_2 \land \lnot P_3 }$ $(\langle (7,0)\rangle ,q_2)$ to make both the individual trajectories synchronized to have same no. of transitions for same transition condition. 
\end{example}

If for a robot, all the nodes in the abstract individual trajectory are  $(*,*)$, then this robot is not doing anything in this team task sequence $C_f$. In such a case, we can directly ignore this robot. For example, consider a $C_f$ $=$ $(\langle (7,0,)(*,*)\rangle ,q_0)$ $\rightarrow$ $(\langle (*,*)(*,*)\rangle ,q_1)$ $\rightarrow$ $(\langle (6,6)(*,*)\rangle ,q_2)$ in which all the coordinates for robot 2 are $(*,*)$.
\begin{example}
Single robot trajectories shown in Figure \ref{exampleshort} (d) and (e) have the same number of nodes and also have the same B\"uchi states. Thus, they are in sync. We can combine them as $\mathcal{R}_c^{suf}$ $=$ $(\langle (6,5),(0,7)\rangle ,q_1)$ $\rightarrow$ $(\langle (6,6),(0,6)\rangle ,q_2)$ $\rightarrow$ $(\langle (6,6),(0,7)\rangle ,q_0)$ $\rightarrow(\langle (6,5),(0,7)\rangle ,q_1)$ with cost equal to sum of individual costs which is $2+2=4$. 
\end{example}
(viii) $\mathtt{Compt\_Prefix}$( ): After synchronizing the individual trajectories, we now know the exact coordinates of the final state $f \in F_r$ in $C_f$. 
For example, $C_f$ in \ref{exampleshort}(a) has $f = (\langle (*,*)(0,7)\rangle ,q_0)$ which we computed as $(\langle (6,6)(0,7)\rangle ,q_0)$. Now, in this step we can find a prefix path from initial state $v_{init} = (s_0,q_0)$ to this computed $f$ using the same steps we used to compute the suffix (we find path instead of cycle). If the suffix has a valid prefix (i.e. the suffix is reachable from the initial state of the robots), we can consider such suffix a valid suffix.
(7) From all those suffix cycles, we choose one with minimum cost and have valid prefix as our final outcome. Project it over $\{ T^1,..,T^n\}$ to obtain $\{ \mathcal{R}^1,..,\mathcal{R}^n\}$.
For the Abstract Reduced Graph $G_r$ shown in Figure \ref{AbstRedGraph}, the suffix cycle with the minimum cost is $C_f = (\langle (6,6),(0,7)\rangle ,q_0) \rightarrow (\langle (6,6),(0,7)\rangle ,q_0)$ with cost 0. The Prefix can be computed accordingly.

\smallskip
\noindent
\textbf{Memoization.}
We can store once generated paths for individual robots and use them later in $\mathtt{Optimal\_Run}$ to reduce the computation time. For example, we can store the actual path computed for the abstract path $(\langle (6,6)\rangle ,q_2)$ $ \xrightarrow[] {\lnot P_2 \land \lnot P_3} $ $ (\langle (*,*)\rangle ,q_0)$ $ \xrightarrow[] {\lnot P_1 \land \lnot P_3\{ N\}} $ $ (\langle (*,*)\rangle ,q_1) $ $ \xrightarrow[]{ P_1 \land \lnot P_2  \land \lnot P_3 \{ \lnot P_1 \land \lnot P_3 \} } $ $(\langle (6,6)\rangle ,q_2) $.

\smallskip
\noindent
\shortversion{
\textbf{Correctness and Optimality.}
Due to the space constraint, we have added the proof of optimality and correctness of \textsf{MT*} to Appendix \ref{AppendixCorrectnessAndOptimality} where we prove that the trajectory generated by \textsf{MT*} satisfies the given LTL formula and minimizes the cost function given by Equation~\eqref{eq3}.
}
\longversion {
\textbf{Correctness and Optimality.}
In this section, we prove the correctness of MT* algorithm. To prove the correctness of \textsf{MT*}, we will have to show that the suffix run which we find in the algorithm satisfies the given LTL formula, and it is the minimum cost suffix run among all the satisfying runs.

\begin{theorem} 
The suffix run $\mathcal{R}_{suf}$ computed by \textsf{MT*} algorithm follows the given LTL formula $\phi$, and it is the minimum cost run among all the $\phi$ satisfying runs.
\end{theorem}
\textbf{\emph{Proof:}} In \textsf{MT*} algorithm, we work on the Abstract Reduced Graph $G_r$, which is a reduced version of the Product Graph $P$. First, we will have to prove that Abstract Reduced Graph preserves all the minimum cost paths starting and ending a state $f_p \in F_P$.
\begin{lemma}
Abstract Reduced Graph preserves all the minimum cost paths starting and ending at a state $f_p \in F_P$.
\end{lemma}
\textbf{\emph{Proof:}} First, we claim that all the final states $F_P$ present in $P$ are preserved in $F_r$. All the incoming transitions to the final B\"uchi automaton states are of type $c_{pos}$. This is because whenever we specify some task in the form of an LTL query, it contains at least one positive proposition(positive propositions represent actual task whereas negative propositions specify constraints to be followed by the robots. And LTL task with only constraints, i.e., negative propositions will be meaningless in the context of robotic applications) and as B\"uchi final state signifies the completion of the given task, the incoming transition to B\"uchi final state will always be of type $c_{pos}$. In $G_r$, we add all the nodes which satisfy $c_{pos}$ transitions, and these nodes are in abstract form.
So, from this, we can say that we add all the nodes from set $F_P$ in abstract form to $G_r$. We denote $F_P$ in abstract form as $F_r$. 
In Abstract Reduced Graph, we add nodes using two conditions. First is the product graph condition, in which added node should be neighbour. In this condition, we added $(*,*)$ as neighbour. We can always substitute any neighbouring node in place of $(*,*)$. So no transition is lost for the nodes added using this condition. The second one is the distant neighbour condition, in which we mean to use $c_{neg}$ type self-loop to establish a path. We use Dijkstra's to establish this path. In distant neighbour condition, we lose some transition as we skip transitions due to $c_{neg}$ self-loop. But, we can use Dijkstra's algorithm with $c{neg}$ constraint and recover the shortest path between the nodes which were added using the distant neighbour condition. So, from these arguments, we can say that Abstract Reduced Graph preserves all the minimum cost paths starting and ending at a state $f_p \in F_P$.
 
Now, consider a simple cycle $C_f$ in Abstract Reduced Graph. It represents a possible task assignment for the robots. We then compute the trajectories for individual robots such that collectively they follow constraints in $C_f$, and each individual robot is computed using either Dijkstra's algorithm or \textsf{A*} algorithm. So, the overall moving cost of the robots is minimized. In \textsf{MT*}, we repeat this procedure for all the possible cycles. So all the possible task assignments are considered.

So, from these arguments, we conclude that the suffix run $\mathcal{R}^{suf}_P$ computed by \textsf{MT*} algorithm follows the given LTL formula $\phi$ and it is the minimum cost run among all the $\phi$ satisfying runs.
}

\smallskip
\noindent
\textbf{Complexity.}
Computation time of MT* increases exponentially with the increase in the number of robots and the size of the LTL specification as these increase the size of the abstract reduced graph and thus the number of cycles to be explored. However, as the size of the abstract reduced graph remains the same with an increase in the size of the workspace, the computation time increases linearly with the increase in the workspace size due to the linear increase in the distance between two locations of interest. This provides a significant advantage over the baseline algorithm. As the precise complexity analysis MT* is complex, we rely on the experimental evaluation to demonstrate its efficacy. 
\section{Evaluation}

\begin{table*}[t]
\begin{center}
\caption{Baseline Solution (B.S.) Vs \textsf{MT*}}
\label{table1}
\small\addtolength{\tabcolsep}{-3pt}
\begin{tabular}{|c|c|c|c|c|c|c|c|c|c|c|c|c|c|c|c|c|}
        \hline
          {}&{}& \multicolumn{7}{c|}{$9 \times 9$ Workspace} & \multicolumn{4}{c|}{$15 \times 15$ Workspace} & \multicolumn{4}{c|}{$30 \times 30$ Workspace}\\
         \cline{3-17}
          {}&{}& \multicolumn{3}{c|}{2 Robots} & \multicolumn{3}{c|}{3 Robots} & 8 Robots & \multicolumn{3}{c|}{2 Robots} & 8 Robots & \multicolumn{3}{c|}{2 Robots} & 8 Robots \\
          \cline{3-17}
         $\phi$ & B\"uchi & B.S. & MT* & Speed & B.S. & MT* & Speed & MT* & B.S. & MT* & Speed & MT* & B.S. & MT* & Speed & MT* \\
           & States & (sec) &(sec) &   Up & (sec) &(sec) &   Up & (sec) & (sec) & (sec) &   Up & (sec) & (sec) & (sec) &   Up & (sec)\\
        \hline
        $\phi_1$  & 12 & 20.6 & 13.6 & \textbf{1.5} & - & 266 & \textbf{-} & 5397.19  & 3635 & 19.8 & \textbf{184} & 5513.59 & - & 37.41 & \textbf{-} & 6022.69 \\
        $\phi_2$  & 5  & 2.8  & 0.1  & \textbf{21.9} & 9786 & 0.61 & \textbf{16037} & 147.11  & 74   & 0.3  & \textbf{243} & 151.54 & 985 & 1.0 & \textbf{985} & 163.47\\
        $\phi_3$  & 5  & 22.3 & 0.9  & \textbf{23.7} & 11665 & 5.5 & \textbf{2107} & 1211.12 & 496  & 2.5  & \textbf{198} & 1196.55 & 9043 & 8.3 & \textbf{1086} & 1311.95\\
        $\phi_4$  & 5  & 1.8  & 0.04 & \textbf{40.0} & 1150 & 0.1 & \textbf{11500} & 66.36  & 38   & 0.06 & \textbf{664} & 66.48 & 728 & 0.12 & \textbf{6129} & 67.53\\
        $\phi_5$  & 5  & 11.4 & 0.05 & \textbf{218.8} & - & 0.19 & \textbf{-} & 382.33 & 579  & 0.07 & \textbf{7846} & 383.66 & 10937 & 0.14 & \textbf{80633} & 383.99\\
        \hline
\end{tabular}
\end{center}
\end{table*}

\begin{table}
  \begin{center}
     \caption{Comparison of No. of Vertices and Edges in Product Graph (P) and Abstract Reduced Graph ($G_r$) for Query $\phi_2$  }
    \label{nodeTable}
  \small\addtolength{\tabcolsep}{-3pt}
    \begin{tabular}{|c|c|c|c|c|c|}
    \hline
    $|\mathcal{W}|$ & $|n|$  & \multicolumn{2}{c|}{P} &  \multicolumn{2}{c|}{$G_r$}\\
    \cline{3-6}
    {}    & {} & $|S_P|$ & $|E_P|$      &  $|S_r|$     & $|E_r|$\\
    \hline
    9x9   & 2 & 9605    & 305767     &     15        &  26  \\
    15x15 & 2 & 152101  & 6329687    &     15        &  26  \\
    30x30 & 2 & 2762245 &  126946183 &     15        &  26  \\
    40x40 & 2 & 9375845 &  442918092 &     15        &  26  \\
    9x9   & 3 & 480254  &  60541878  &     19        &  66  \\
    \hline
\end{tabular} 
  \end{center}
\end{table}


\begin{figure} 
\subfigure[Trajectories for $\phi_2$]{\centering \label{traj_short}\includegraphics[height=22mm,width=25mm]{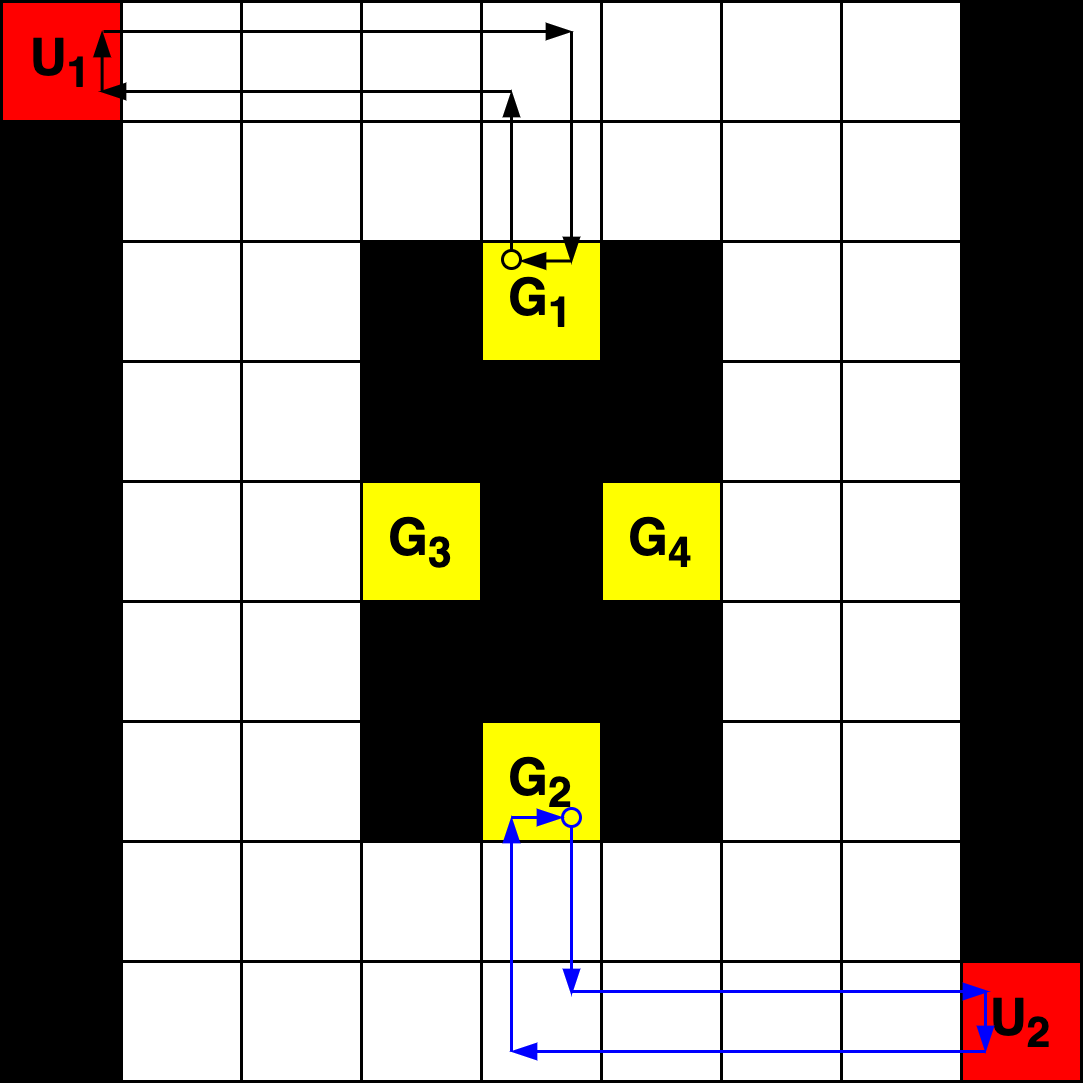}}
\subfigure[Computation Time(s)  Vs Workspace Size  ]{ \label{g2}\includegraphics[height=22mm,width=30mm]{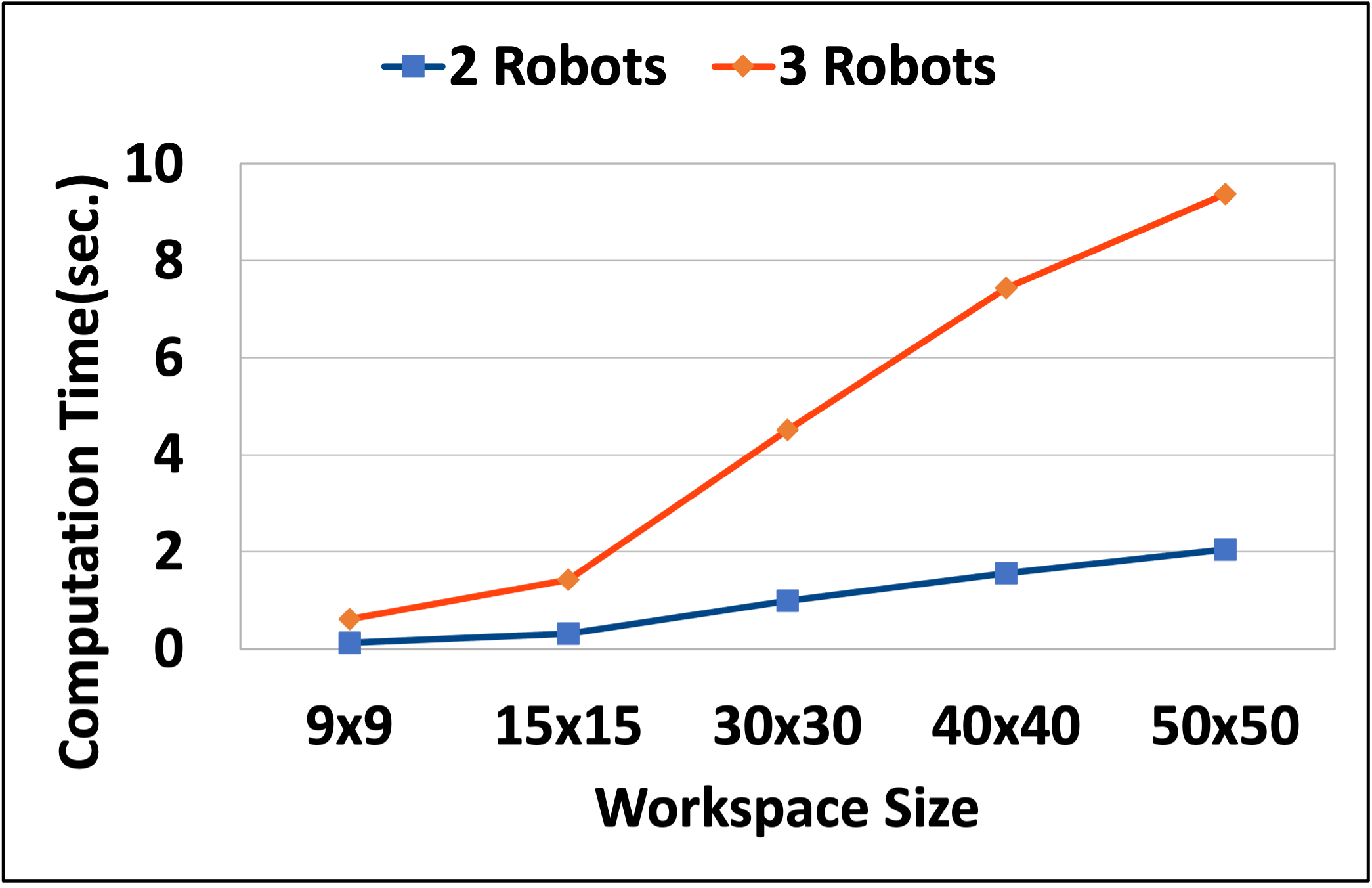}}
\subfigure[Computation Time(s) Vs No. of Robots]{\centering \label{g1}\includegraphics[height=22mm,width=30mm]{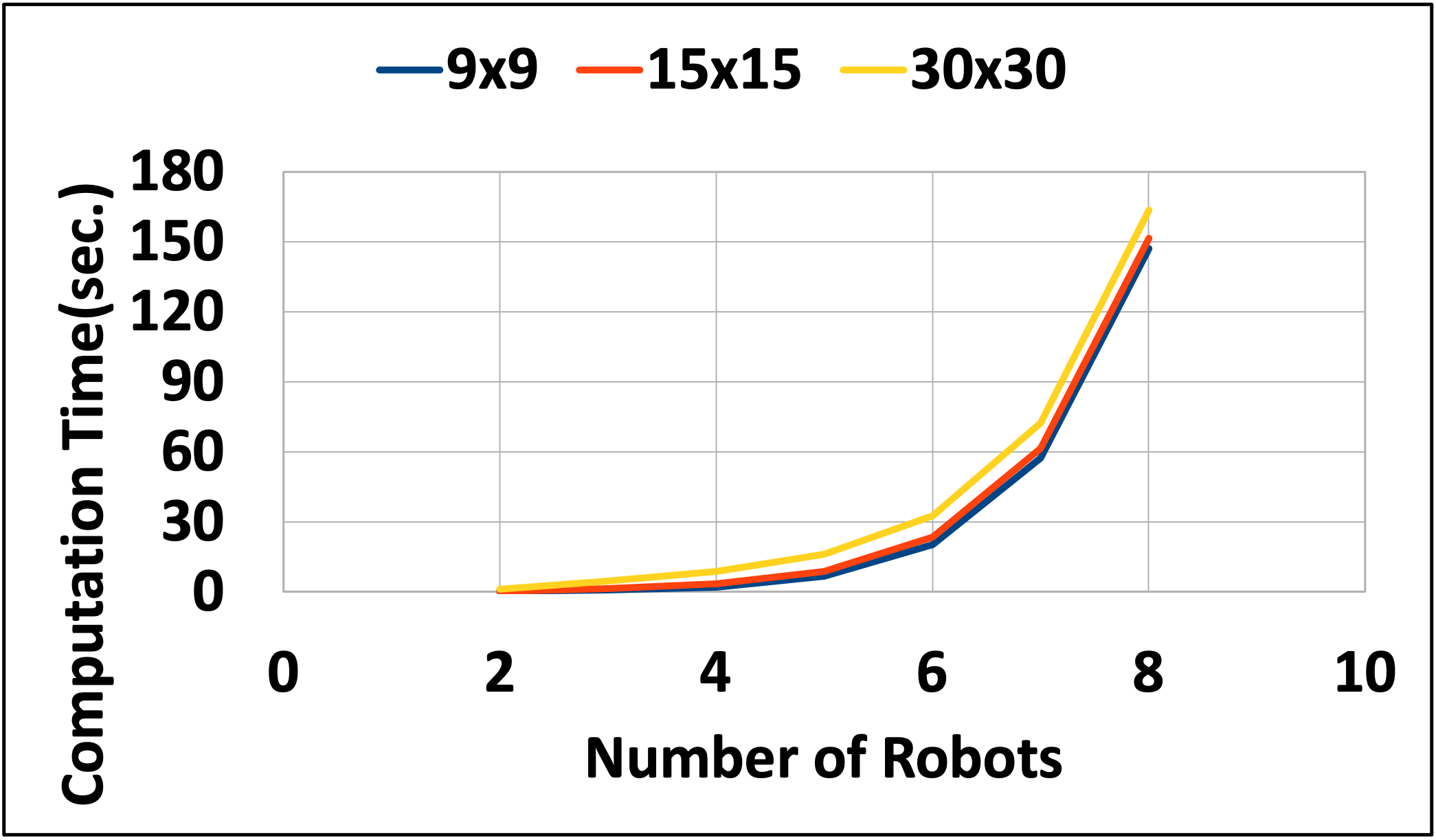}}
\caption{Results for LTL Query $\phi_2$}
\label{g}
\end{figure}


In this section, we present several results to establish the computational efficiency of \textsf{MT*} algorithm against the baseline solution~\cite{UlusoySDBR13}. The results have been obtained on a desktop computer with a $3.4$\,GHz quadcore processor with $16$\,GB of RAM. We use \textsf{LTL2TGBA} tool~\cite{DuretLutz2004SPOTAE} as the LTL query to B\"{u}chi automaton converter. The C++ implementations of \textsf{MT*} and Baseline algorithms and a simulation video are submitted as supplementary materials.
We use the $2$-D workspace as shown in Figure~\ref{traj_short} (borrowed from~\cite{UlusoySDBR13}). Each grid-cell has 4 neighbours. The cost of each edge between the neighbouring cells is $1$ unit. In the workspace, $U_1 \text{ and } U_2$ are data upload locations 1 and 2, whereas $G_1, G_2, G_3 \text{ and } G_4$ are data gather locations $1$ to $4$.


We have evaluated \textsf{MT*} algorithm for five LTL queries $\phi_{1}, \phi_{2}, \ldots, \phi_{5}$ borrowed from~\cite{UlusoySDBR13}. We define propositions over the workspace in the following way: $gather$: Data has been gathered from a gather station, $rXgather$: Robot X has gathered data from a gather station,
$gatherY$: Data has been gathered from the gather station Y,
$rXgatherY$: Robot X has gathered data from the gather station Y. We define propositions for `upload' in the same way.

\shortversion{
Due to space constraints, we discuss only one of the five queries in this section.
Detailed results for all the five queries are available in Appendix \ref{AppendixUseCases}.



\noindent 
\textbf{Query $\phi_2$}: The mission is ``Each Robot must repeatedly visit a data gather location at same time synchronously to gather data and then upload that data to an upload station before gathering the new data again.''
{ \setlength{\mathindent}{5pt}
\begin{align} \nonumber
    \phi_2 =\;&  \square \Diamond gather \land \square(gather \Rightarrow (r1gather \land r2gather)) \; \land \\
    \nonumber\ \;&\square(r1gather \Rightarrow \text{X} ( \lnot (r1gather) \text{U} (r1upload) ) ) \; \land \\
    \nonumber\ \;&\square ( r2gather \Rightarrow \text{X}( \lnot (r2gather) \text{U} (r2upload)))
\end{align}}
\noindent The generated trajectory for query $\phi_2$ is shown in Figure \ref{traj_short} in which the hollow circle represents the start of the trajectory.  For $\phi_2$, Robot $1$ gathers from $G1$ and Robot $2$ from $G2$ at the same time to induce the synchronization constraint $\square(gather \Rightarrow (r1gather \land r2gather))$. They upload it to the nearest upload station to minimize the total cost and move again to gather data at $G1$ and $G2$ respectively to complete the cyclic trajectory. The total cost of the trajectory for both the robots is $24$.
} 

\longversion {

\begin{figure*}[!h]
\subfigure[Trajectory for Query $\phi_1$]{\centering \label{traj1_full}\includegraphics[width=33mm]{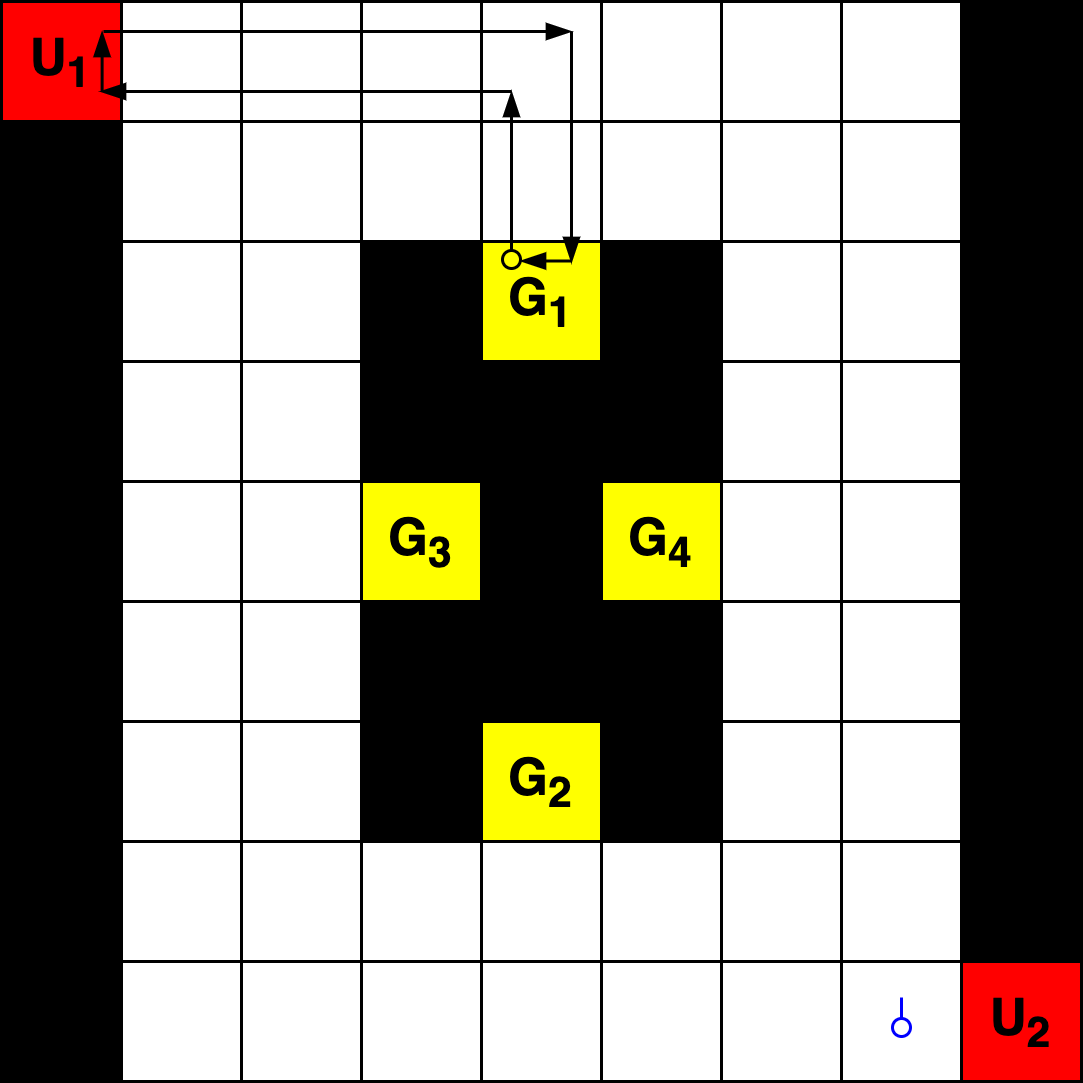}}
\hspace{2mm}
\subfigure[Trajectory for Query $\phi_2$]{\centering \label{traj2_full}\includegraphics[width=32mm]{Images/Example2.png}}
\hspace{2mm}
\subfigure[Trajectory for Query $\phi_3$]{\centering \label{traj3_full}\includegraphics[width=33mm]{Images/Example2.png}}
\hspace{2mm}
\subfigure[Trajectory for Query $\phi_4$]{\centering \label{traj4_full}\includegraphics[width=33mm]{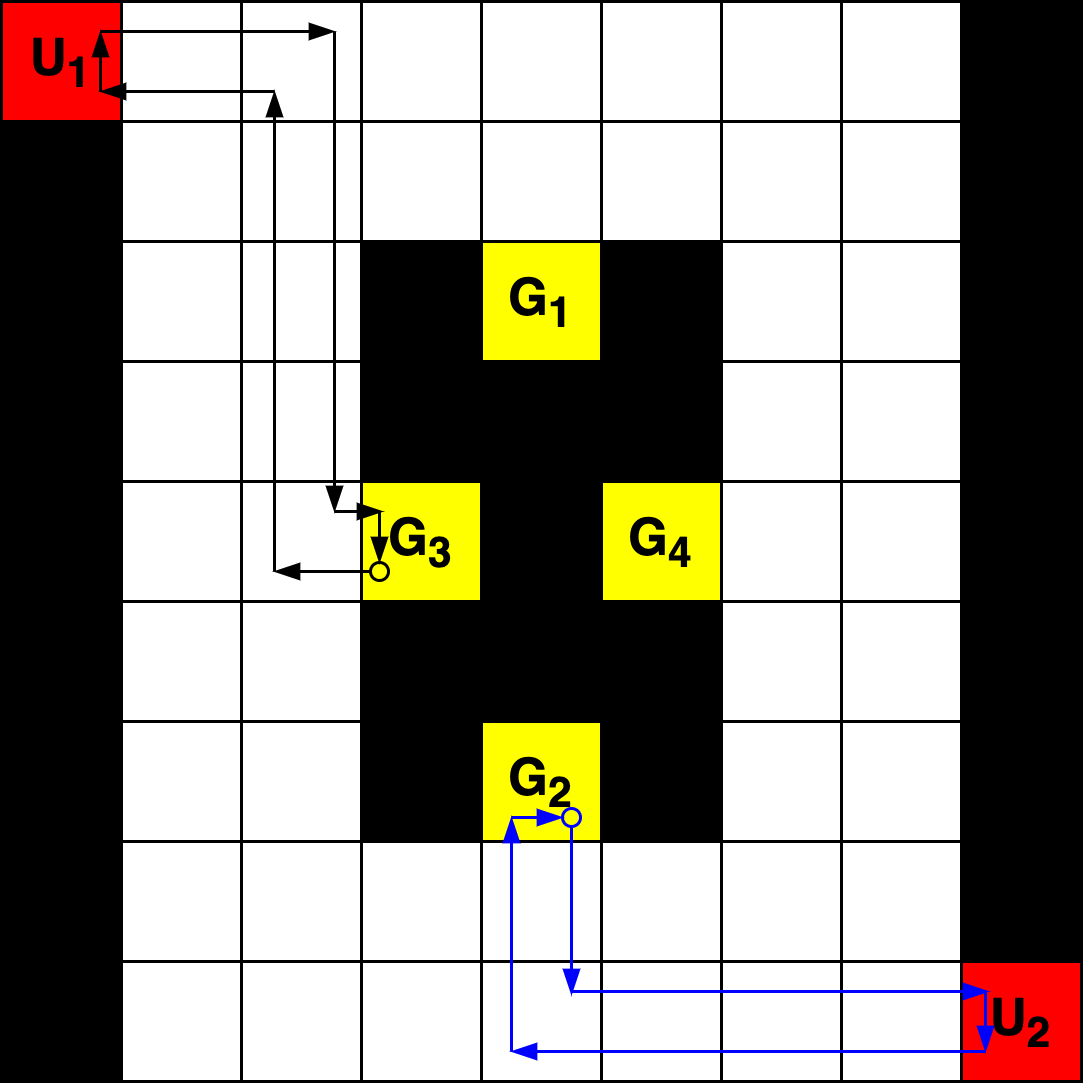}}
\hspace{2mm}
\subfigure[Trajectory for Query $\phi_5$]{\centering \label{traj5_full}\includegraphics[width=33mm]{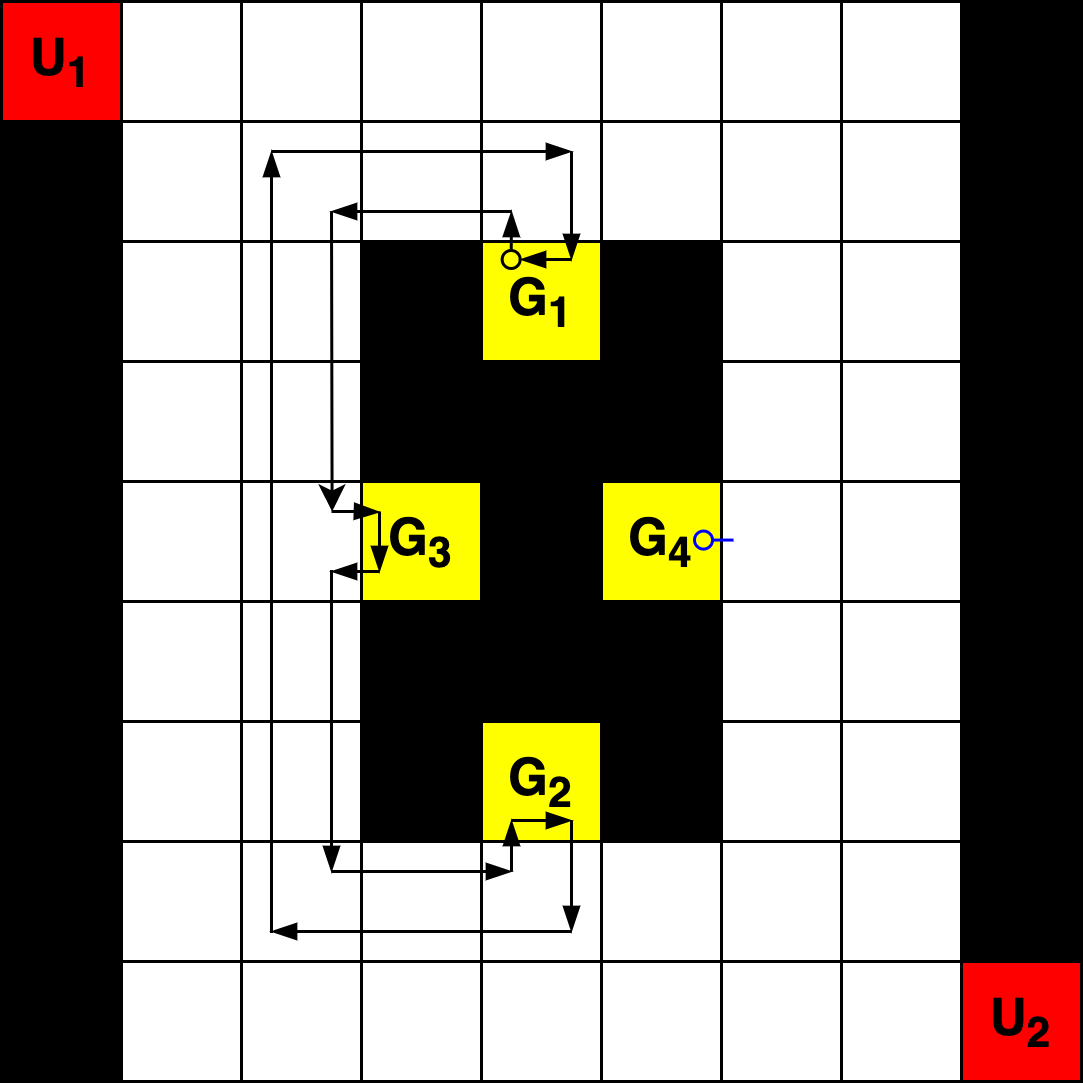}}
\caption{Generated Trajectories for a Two Robot System}
\label{traj_full}
\end{figure*}


\noindent \textbf{Query $\phi_1$}: The mission is ``Repeatedly gather data from data gather locations and once you gather the data, upload it to data upload location before gathering new data.''
\setlength{\mathindent}{5pt}{
\begin{align} \nonumber
    \phi_1 =\;&  \square \Diamond gather \, \land \\
    \nonumber\ \;&\square(r1gather \implies \text{X} ( \lnot (r1gather) \, \text{U} \, (r1upload)  )) \, \land \\
    \nonumber\ \;&\square ( r2gather \implies \text{X}( \lnot (r2gather) \text{U} (r2upload))) 
\end{align}}
This task induces a sequence among the locations to be visited. Here, Once a robot gathers data at any data gathering station, it must visit a data upload station before it can gather the data again. It also induces a response(visit upload station) to an event(visit to gather station) in the robot trajectory. The trajectory generated by this specification is shown in \ref{traj_full}(a). We can see that only robot $1$ circling between $G1$ and $U1$ and robot $2$ remains at its initial location. This happened because we never said in the specification that both the robots should gather data. We only asked to gather data, and a trajectory is generated based on that. Robot $2$ remains at its location, as the cost of not moving is $0$, and we are trying to minimize the cost. The cost of this trajectory is $12$ (number of movements).

\noindent \textbf{Query $\phi_2$}: The mission is ``Each Robot must repeatedly visit a data gather location at same time synchronously to gather data and then upload that data to an upload station before gathering the new data again.''
\setlength{\mathindent}{5pt}{
\begin{align} \nonumber
    \phi_2 =\;&  \square \Diamond gather \, \land \\
    \nonumber\ \;&\square(r1gather \implies \text{X} ( \lnot (r1gather) \,\text{U}\, (r1upload ) )) \, \land \\
    \nonumber\ \;&\square ( r2gather \implies \text{X}( \lnot (r2gather) \,\text{U}\, (r2upload))) \, \land \\
    \nonumber\ \;&\square(gather \implies (r1gather \land r2gather))
\end{align}}
The newly added constraint asks both the robots to gather the data and that too at the same time. These kinds of constraints can be used to induce synchronization among the robots.  The trajectories for this specification are shown in Figure \ref{traj_full}(b). Here, both the robots gather data at the same time and upload it before visiting the data gather station again. In Figure \ref{traj_full}(b), a small circle on the robot trajectory denotes the starting point of the cycle. The cost of the trajectories generated for this specification is $24$. 

\noindent \textbf{Query $\phi_3$}: The mission is ``Each Robot must repeatedly visit data gather location at same time synchronously to gather data but not to the same gather station and upload that data at upload station before gathering the data again.''
\setlength{\mathindent}{5pt}{
\begin{align} \nonumber
    \phi_3 =\;&  \phi_2 \; \land \; \square(\lnot ( r1gather1 \land r2gather1) \; \land \\
    \nonumber\ \;& \lnot ( r1gather2 \land r2gather2)\; \land \\
    \nonumber\ \;& \lnot ( r1gather3 \land r2gather3) \; \land \\ \nonumber\ \;& \lnot ( r1gather4 \land r2gather4))
\end{align}}
In case 2, the robots were asked to gather data at the same time. One of the possible trajectories is shown in Figure \ref{traj_full}(c), in which both the robots gather data from the same gather station and upload it to the same upload station. In case 3, we ask the robots to gather data from the different gather stations. So, Robot 1 gathers data from $G1$ and Robot $2$ from $G2$. This way, they follow the specification and also minimize the cost of the movement. This kind of specification can be used to avoid repetitive work. The trajectories generated for this specification are shown in Figure \ref{traj_full}(c). We can also observe that the data gathering task has been synchronized but at different data gathering locations. The cost for this run is $24$. 

\noindent \textbf{Query $\phi_4$}: The mission is ``Robot $1$ must repeatedly visit data gather location $G3$ and robot $2$ must repeatedly visit data gather location $G2$ at the same time to gather data and upload that data at upload station before gathering the data again.''
\setlength{\mathindent}{5pt}{
\begin{align} \nonumber
    \phi_4 =\;&  \square \Diamond gather \; \land \\
    \nonumber\ \;&\square(r1gather \implies \text{X} ( \lnot (r1gather) \,\text{U}\, (r1upload) ) )) \; \land \\
    \nonumber\ \;&\square ( r2gather \implies \text{X}( \lnot (r2gather) \,\text{U}\, (r2upload))) \; \land \\
    \nonumber\ \;&\square(gather \implies ((r1gather3) \land (r2gather2)))
\end{align}}
In this specification, we specifically assign the data gather location to each of the robots, and they have to choose the closest data upload station to upload that data before gathering more data. The trajectories for both the robots are shown in Figure~\ref{traj_full}(d). The total cost of the run is $26$. This kind of specification can be used to assign a certain task to a specific robot from a team of robots.

\noindent \textbf{Query $\phi_5$}: The mission is ``Gather data from all the gather stations.''
\setlength{\mathindent}{5pt}{
\begin{align} \nonumber
    \phi_5 =\;  &\square \Diamond gather1 \ \land \ \square \Diamond gather2 \ \land \ \square \Diamond gather3 \ \land  \ \\
    \nonumber \ \; &\square \Diamond gather4
\end{align}}
In this case, we are only interested in gathering data. But data must be gathered from all the gather stations. Here, we just repeatedly want to gather data from all the gather stations. The robot trajectories are shown in Figure~\ref{traj_full}(e). Here, robot $1$ is covering three of the four gather stations, and robot $2$ just stays at the gather station $4$. This happened because this kind of trajectory has the minimum cost, i.e., this is the best possible task assignment possible. The cost of this run is $26$.
}

In Table~\ref{table1}, we list down the computation times of the baseline solution and \textsf{MT*}, and the spped-up achieved by \textsf{MT*} over the baseline solution for queries $\phi_1,...,\phi_5$. B\"uchi states column lists the number of states in the B\"uchi automaton for the corresponding LTL mission. 
We list these results for different sizes of workspaces. A $9 \times 9$ workspace is shown in Figure~\ref{traj_short}. The $15 \times 15$ and $30 \times 30$ workspaces are similar to the $9 \times 9$ map with the same number of data gather and data upload locations. In the table, we can observe significant speed up that \textsf{MT*} achieves over the baseline solution. We have shown `-' for the entries which we could not compute due to insufficient RAM ($16$\,GB) or very high computation time ($>10000\,\si\sec$). For $8$ robots, we have only shown computation time for \textsf{MT*} as we could not generate results for baseline solution beyond $3$ robots due to very high computation time and memory requirement.


For the graphs in Figure~\ref{g}, we have used workspaces from size $9\times9$ till $50\times50$.  
In Figure~\ref{g2}, we observe the performance of \textsf{MT*} with the increase in workspace size for $2$ and $3$ robot systems for query $\phi_2$.  
The computation time of \textsf{MT*} increases almost linearly for all the LTL queries. 
This is because the size of the abstract reduced graph remains the same with the increase in the workspace size. The computation time for single robot trajectories in the procedure $\mathtt{Optimal\_Run}$  increases with the increase in the workspace size, and thus \textsf{MT*} achieves a linear increase in the computation time with the increase in the workspace size. These results are consistent for other queries and workspaces. 
In Figure \ref{g1}, we observe that computation time of \textsf{MT*} increases exponentially with the increase in the number of robots for all the LTL queries. The graph shown is for query $\phi_2$. This is because, with the increase in the number of robots, the number of possible task assignments increases exponentially.

In Table~\ref{nodeTable}, we compare the number of vertices and edges in Product Graph $P$ with Abstract Reduced Graph $G_r$ for different workspace sizes $|\mathcal{W}|$ and different number of robots $|n|$. 
The Abstract Reduced graph remains the same with the increase in the workspace size and is significantly smaller than the product graph, and that is why it has superior performance over the baseline solution in terms of computation time. 
The sizes of the graphs are also a direct indicator of the memory requirement of the algorithms.

\section{Discussions}


Our proposed algorithm \textsf{MT*} is substantially faster than the state-of-the-art algorithm to solve the multi-robot LTL optimal path planning problem.
As our experimental results establish, the computation time for \textsf{MT*} increases linearly with the increase in the size of the workspace. We were able to generate plans for up to $8$ robots for $30\times30$ sized workspace, whereas the state-of-the-art algorithm hardly scales up to $3$ robots over $15\times15$ sized workspace. 

In \textsf{MT*}, we evaluate the cost of the cycles containing a final state one by one while keeping track of the minimum cost cycle. Once we are done with the computation of the first cycle, we have a valid solution (a trajectory satisfying the LTL specification). In the subsequent iterations, we look for a more optimal solution. Thus, MT* is a good candidate for an \emph{anytime} implementation (like anytime \textsf{A*}~\cite{LikhachevGT03}). Moreover, it is possible to parallelize the evaluation of the suffix cycles (for different final states) in \textsf{MT*} to boost the performance further. 
 
\textsf{MT*} does not attempts to provide collision-free trajectories unless the requirement of collision avoidance is explicitly specified in the input LTL formula.
Thus, \textsf{MT*} is useful for high-level strategic planning for a temporal logic specification. We assume that, during the actual execution of the plans, some dynamic real-time collision avoidance algorithms such as the ones presented in~\cite{BergLM08,BergGLM09,HennesCMT12,BestNM16,LongFLLZP18} will be employed to ensure collision avoidance among the robots.

In our future work, we plan to extend our algorithm to deal with dynamic obstacles~\cite{KoenigL02,Cannon2012RealTimeMP} and preferential constraints~\cite{BaierM08,BaierBM09}. We also plan to explore the possibility of applying the recently developed heuristics for multi-agent path finding~\cite{LiFB0K19} to make MT* more scalable.



\bibliographystyle{abbrv}
\bibliography{bb}

\shortversion {
\appendix

\subsection{Baseline Solution Approach}\label{AppendixBaselineApproach}
The baseline solution to above problem uses the automata-theoretic model checking approach~\cite{UlusoySDBR13}, the steps of which are outlined in the Algorithm~\ref{baselineSolution}.

\begin{algorithm2e}
\SetAlgoLined
\DontPrintSemicolon
\setstretch{0.9}
\textbf{Input:} Transition systems $\{ T^1,...,T^n\}$, $\phi$: An LTL formula\;
\textbf{Output: }A set of runs $\{ \mathcal{R}^1,...,\mathcal{R}^n \}$that satisfies $\phi$\;
\BlankLine
Construct a joint transition system $T$\;
Convert $\phi$ to a B\"uchi automaton $B$\;
Compute the product automaton $P = T \times B$\;
\For{all $f \in F_P$ }{
$\mathcal{R}_f^{suf} \gets \mathtt{Dijkstra's\_Algorithm}(\: \emph{f} \:,\:  \emph{f}\:)$\;
$\mathcal{R}_f^{pre} \gets \mathtt{Dijkstra's\_Algorithm}(\: S_{P,0} \:,\:  \emph{f}\:)$
}
$\mathcal{R}_P^{suf} \gets$ minimum of all $\mathcal{R}_f^{suf}$ \;
$\mathcal{R}_P^{pre} \gets$ prefix of $\mathcal{R}_P^{suf}$ \;
$\mathcal{R}_P = \mathcal{R}_P^{suf} . \mathcal{R}_P^{pre} $\;
Project $\mathcal{R}_P$ over $T$ to compute $\mathcal{R}_T$\;
Project $\mathcal{R}_T$ over $\{ T^1,...,T^n\}$ to obtain runs $\{\mathcal{R}^1,...,\mathcal{R}^n\}$
\caption{Baseline\_Solution}
\label{baselineSolution}
\end{algorithm2e} 

The first step in this algorithm is to compute
the joint transition system $T$ from the transition systems of the individual robots $T^i$. Then we compute B\"uchi automaton from the given LTL query $\phi$. We then compute the product automaton of $T$ and $B$. In this product automaton, for each final state $f \in F_P$, we find a prefix run starting from initial state $S_{P,0}$ to $f \in F_P$ and then find minimum cost cycle starting and ending at $f$ using Dijkstra's algorithm. We then choose the prefix-suffix pair with the smallest $\mathcal{C}(\mathcal{R}_P)$ cost i.e. the pair with smallest suffix cost, and project it on $T$ to obtain the run $\mathcal{R}_T$ which represents the joint motion all the robots in $\mathcal{W}$. We then  project $\mathcal{R}_T$ over individual $T^i$ to obtain the  of the individual robots $\mathcal{R}^i$. The trajectory $\mathcal{R}^i$ for the $i$-th robot provides us with the cyclic trajectory which the robot can follow repetitively to complete the given task $\phi$ repetitively.  

In the following section, we present MT$^*$ algorithm that provides a significantly improved running time for generating an optimal trajectory satisfying a given LTL query.

\begin{figure*}[!h]
\subfigure[Trajectory for Query $\phi_1$]{\centering \label{traj1_full}\includegraphics[width=33mm]{Images/Example 1.png}}
\hspace{2mm}
\subfigure[Trajectory for Query $\phi_2$]{\centering \label{traj2_full}\includegraphics[width=32mm]{Images/Example 2.png}}
\hspace{2mm}
\subfigure[Trajectory for Query $\phi_3$]{\centering \label{traj3_full}\includegraphics[width=33mm]{Images/Example 2.png}}
\hspace{2mm}
\subfigure[Trajectory for Query $\phi_4$]{\centering \label{traj4_full}\includegraphics[width=33mm]{Images/Example 4.png}}
\hspace{2mm}
\subfigure[Trajectory for Query $\phi_5$]{\centering \label{traj5_full}\includegraphics[width=33mm]{Images/Example 5.png}}
\caption{Generated Trajectories for a Two Robot System}
\label{traj_full}
\end{figure*}

\subsection{Correctness and Optimality}\label{AppendixCorrectnessAndOptimality}

In this section, we prove the correctness of MT* algorithm. To prove the correctness of \textsf{MT*}, we will have to show that the suffix run which we find in the algorithm satisfies the given LTL formula and it is the minimum cost suffix run among all the satisfying runs.

\begin{theorem} 
The suffix run $\mathcal{R}_{suf}$ computed by \textsf{MT*} algorithm follows the given LTL formula $\phi$ and it is the minimum cost run among all the $\phi$ satisfying runs.
\end{theorem}
\textbf{\emph{Proof:}} In \textsf{MT*} algorithm, we work on the Abstract Reduced Graph $G_r$ which is reduced version of the Product Graph $P$. So, first We will have to prove that Abstract Reduced Graph preserves all the minimum cost paths starting and ending a state $f_p \in F_P$.
\begin{lemma}
Abstract Reduced Graph preserves all the minimum cost paths starting and ending at a state $f_p \in F_P$.
\end{lemma}
\textbf{\emph{Proof:}} First, we claim that all the final states $F_P$ present in $P$ are preserved in $F_r$. All the incoming transitions to the final B\"uchi automaton states are of type $c_{pos}$. This is because, whenever we specify some task in the form of an LTL query, it contains at least one positive proposition(positive propositions represent actual task whereas negative propositions specify constraints to be followed by the robots. And LTL task with only constraints i.e. negative propositions will be meaningless in the context of robotic applications) and as B\"uchi final state signifies the completion of the given task, incoming transition to B\"uchi final state will always be of type $c_{pos}$. In $G_r$, we add all the nodes which satisfy $c_{pos}$ transitions and these nodes are in abstract form.
So, from this, we can say that, we add all the nodes from set $F_P$ in abstract form to $G_r$. We denote $F_P$ in abstract form as $F_r$. 
In Abstract Reduced Graph, we add nodes using two conditions. First is product graph condition, in which added node should be neighbour. In this condition, we added $(*,*)$ as neighbour. We can always substitute any neighbouring node in place of $(*,*)$. So no transition is lost for the nodes added using this condition. Second one is distant neighbour condition, in which we mean to use $c_{neg}$ type self loop to establish a path. We use Dijkstra's to establish this path. In distant neighbour condition, we loose some transition as we skip transitions due to $c_{neg}$ self loop. But, we can use Dijkstra's algorithm with $c{neg}$ constraint and recover shortest path between the nodes which were added using distant neighbour condition. So, from these arguments, we can say that Abstract Reduced Graph preserves all the minimum cost paths starting and ending at a state $f_p \in F_P$.
 
Now, consider a simple cycle $C_f$ in Abstract Reduced Graph. It represents a possible task assignment for the robots. We then compute the trajectories for individual robots such that collectively they follow constraints in $C_f$ and each individual robot is computed using either Dijkstra's algorithm or \textsf{A*} algorithm. So, the overall moving cost of the robots is minimised. In \textsf{MT*}, we repeat this procedure for all the possible cycles. So all the possible task assignments are considered.

So, from these arguments we conclude that the suffix run $\mathcal{R}^{suf}_P$ computed by \textsf{MT*} algorithm follows the given LTL formula $\phi$ and it is the minimum cost run among all the $\phi$ satisfying runs.

\subsection{Details of the Use-Cases}\label{AppendixUseCases}

\noindent \textbf{Query $\phi_1$}: The mission is ``Repeatedly gather data from data gather locations and once you gather the data, upload it to data upload location before gathering new data.''
\setlength{\mathindent}{5pt}{
\begin{align*} \nonumber
    \phi_1 =\;&  \square \Diamond gather \; \land \\
    \nonumber\ \;&\square(r1gather \implies \text{X} ( \lnot (r1gather) \, \text{U} \, (r1upload)  )) \; \land \\
    \nonumber\ \;&\square ( r2gather \implies \text{X}( \lnot (r2gather) \, \text{U} \, (r2upload))) 
\end{align*}}
This task induces a sequence among the locations to be visited. Here, Once a robot gathers data at any data gathering station, it must visit a data upload station before it can gather the data again. It also induces a response(visit upload station) to an event(visit to gather station) in the robot trajectory. The trajectory generated by this specification is shown in \ref{traj_full}(a). We can see that, only robot 1 circling between $G1$ and $U1$ and robot $2$ remains at it's initial location. This happened because, we never said in the specification that both the robots should gather data. We only asked to gather data and trajectory is generated based on that. Robot $2$ remains at it's location, as cost of not moving is $0$ and we are trying to minimize the cost. Cost of this trajectory is $12$ (number of movements).

\noindent \textbf{Query $\phi_2$}: The mission is ``Each Robot must repeatedly visit a data gather location at same time synchronously to gather data and then upload that data to an upload station before gathering the new data again.''
\setlength{\mathindent}{5pt}{
\begin{align} \nonumber 
    \phi_2 =\;&  \square \Diamond gather \; \land \\
    \nonumber\ \;&\square(r1gather \implies \text{X} ( \lnot (r1gather) \, \text{U} \, (r1upload ) )) \; \land \\
    \nonumber\ \;&\square ( r2gather \implies \text{X}( \lnot (r2gather) \, \text{U} \, (r2upload))) \; \land \\
    \nonumber\ \;&\square(gather \implies (r1gather \land r2gather))
\end{align}}
New added constraint asks both the robots to gather the data and that to at the same time. These kind of constraints can be used to induce the synchronization among the robots.  The trajectories for this specification are shown in Figure \ref{traj_full}(b). Here, both the robots gather data at the same time and upload it before visiting the data gather station again. In Figure \ref{traj_full}(b), small circle on the robot trajectory denotes the starting point of the cycle. The cost of the trajectories generated for this specification is $24$. 

\noindent \textbf{Query $\phi_3$}: The mission is ``Each Robot must repeatedly visit data gather location at same time synchronously to gather data but not to the same gather station and upload that data at upload station before gathering the data again.''
\setlength{\mathindent}{5pt}{
\begin{align} \nonumber
    \phi_3 =\;&  \phi_2 \; \land \; \square(\lnot ( r1gather1 \land r2gather1) \; \land \\
    \nonumber\ \;& \lnot ( r1gather2 \land r2gather2)\; \land \\
    \nonumber\ \;& \lnot ( r1gather3 \land r2gather3)\; \land \\ \nonumber\ \;& \lnot ( r1gather4 \land r2gather4))
\end{align}}
In case 2, Robots were asked to gather data at a same time. One of the possible trajectories is shown in Figure \ref{traj_full}(c), in which both the robots gather data from the same gather station and upload it to the same upload station. In case 3, we ask the robots to gather data from the different gather stations. So, Robot 1 gathers data from $G1$ and Robot $2$ from $G2$. This way, they follow the specification and also minimize the cost of the movement. This kind of specification can be used to avoid repetitive work. The trajectories generated for this specification is shown in Figure \ref{traj_full}(c). We can also observe that, data gathering task has been synchronized but at different data gather locations. The cost for this run is $24$. 

\noindent \textbf{Query $\phi_4$}: The mission is ``Robot 1 must repeatedly visit data gather location $G3$ and robot $2$ must repeatedly visit data gather location $G2$ at the same time to gather data and upload that data at upload station before gathering the data again.''
\setlength{\mathindent}{5pt}{
\begin{align} \nonumber
    \phi_4 =\;&  \square \Diamond gather \; \land \\
    \nonumber\ \;&\square(r1gather \implies \text{X} ( \lnot (r1gather) \, \text{U} \, (r1upload) ) )) \;  \land \\
    \nonumber\ \;&\square ( r2gather \implies \text{X}( \lnot (r2gather) \, \text{U} \, (r2upload))) \; \land \\
    \nonumber\ \;&\square(gather \implies ((r1gather3) \; \land (r2gather2)))
\end{align}}
In this specification, we specifically assign the data gather location to each of the robots and they have to choose the closest data upload station to upload that data before gathering more data. The trajectories for both the robots are shown in Figure~\ref{traj_full}(d). The total cost of the run is $26$. This kind of specification can be used to  assign certain task to a specific robot from a team of robots.

\noindent \textbf{Query $\phi_5$}: The mission is ``Gather data from all the gather stations.''
\setlength{\mathindent}{5pt}{
\begin{align}  \nonumber
    \phi_5 =\;  &\square \Diamond gather1 \ \; \land \ \square \Diamond gather2 \ \land \ \square \Diamond gather3 \; \land  \ \\
    \nonumber \; &\square \Diamond gather4
\end{align}}
In this case, we are only interested in gathering data. But data must be gathered from all the gather stations. Here, we just repeatedly want to gather data from all the gather stations. The robot trajectories are shown in Figure~\ref{traj_full}(e). Here, robot 1 is covering three of the four gather stations and robot $2$ just stays at gather station $4$. This happened because this is kind of trajectory has minimum cost, i.e., this is best possible task assignment possible. The cost of this run is $26$.

}

\end{document}